\documentclass[11pt]{article}

\usepackage{url}
\usepackage[usenames,dvipsnames,svgnames,table]{xcolor}
\usepackage{color}
\usepackage{graphicx}
\usepackage{amssymb}
\usepackage{amsmath,amsthm,paralist}
\usepackage{amsbsy}
\usepackage{cite}
\usepackage{algorithm,algorithmic}
\usepackage[margin=1in]{geometry}
\usepackage{afterpage}
\usepackage{multirow}
\usepackage{caption}
\usepackage{bbm}
\usepackage{mathtools}

\usepackage{epstopdf}
\usepackage{xspace} 

\usepackage{comment}

\usepackage{hyperref}
\hypersetup{
    colorlinks=true, 
    linktoc=all,     
    linkcolor=blue,  
}

\allowdisplaybreaks

\newtheorem{cor}{Corollary}

\newtheorem{theorem}{Theorem}

\theoremstyle{remark}
\newtheorem{remark}{Remark}

\newcommand{\bitem}{\begin{itemize}}
\newcommand{\eitem}{\end{itemize}}

\newcommand{\supp}{\mathrm{supp}}
\newcommand{\beqn}{\begin{equation}}
\newcommand{\eeqn}{\end{equation}}
\newcommand{\balign}{\begin{align}}
\newcommand{\ealign}{\end{align}}

\newcommand{\RF}{membership index\xspace}

\def \R {\mathbb{R}}

\DeclareMathOperator*{\argmax}{arg max}

\DeclareMathOperator*{\sign}{sign}
\newcommand{\GamLi}[2]{\Lambda_{#1,#2}}  
\makeatletter
\renewcommand\paragraph{\@startsection{paragraph}{4}{\z@}%
            {-2.5ex\@plus -1ex \@minus -.25ex}%
            {1.25ex \@plus .25ex}%
            {\normalfont\normalsize\bfseries}}
\makeatother
\definecolor{orange}{rgb}{1,0.5,0}

\begin{document}

\title{Simple Classification using Binary Data}

\author{Deanna Needell, Rayan Saab, Tina Woolf}

\maketitle

\begin{abstract}
Binary, or one-bit, representations of data arise naturally in many applications, and are appealing in both hardware implementations and algorithm design. In this work, we study the problem of data classification from binary data and propose a framework with low computation and resource costs. We illustrate the utility of the proposed approach through stylized and realistic numerical experiments, and provide a theoretical analysis for a simple case. We hope that our framework and analysis will serve as a foundation for studying similar types of approaches. 
\end{abstract}

\section{Introduction}
Our focus is on data classification problems in which only a \textit{binary} representation of the data is available. Such binary representations may arise under a variety of circumstances. In some cases, they may arise naturally due to compressive acquisition. For example, distributed systems may have bandwidth and energy constraints that necessitate extremely coarse quantization of the measurements \cite{fang2014sparse}. A binary data representation can also be particularly appealing in hardware implementations because it is inexpensive to compute and promotes a fast hardware device \cite{JacquLBB_Robust,LaskaWYB_Trust}; such benefits have contributed to the success, for example, of 1-bit Sigma-Delta converters \cite{aziz1996overview,candy1962oversampling}. Alternatively, binary, heavily quantized, or compressed representations may be part of the classification algorithm design in the interest of data compression and speed (see, e.g., \cite{BoufoB_1Bit,hunter2010compressive,gupta2010sample,hahn2014adaptive}). The goal of this paper is to present a framework for performing learning inferences, such as classification, from highly quantized data representations -- we focus on the extreme case of 1-bit (binary) representations. Let us begin with the mathematical formulation of this problem. 

{\bfseries Problem Formulation.} Let $\{x_i\}_{i=1}^{p}\subset \R^n$ be a point cloud represented via a matrix $$X = [x_1\,\, x_2\,\, \cdots \,\, x_p] \in \R^{n\times p}.$$ Moreover,   let $A: \R^n \to \R^m$ be a linear map, and  denote by $\sign: \R \to \R$ the sign operator given by 
\begin{align*}
\sign(a) = 
\begin{cases}
1 & a\geq 0 \\
-1 & a<0.
\end{cases}
\end{align*}
Without risk of confusion, we overload the above notation so the sign operator can apply to matrices (entrywise). In particular,  for an
$m$ by $p$ matrix $M$, and $(i,j) \in [m]\times [p]$, we define $\sign(M)$ as the $m\times p$ matrix with entries $$(\sign(M))_{i,j} := \sign(M_{i,j}).$$

We consider the setting where a classification algorithm has access to training data of the form $Q=\sign(AX)$,  along with a vector of associated labels 
  $b = (b_1, \,\, \cdots \,\, , b_p )\in\{1,\dots,G\}^p$, indicating the membership of each $x_i$ to exactly one of $G$ classes. Here, $A$ is an $m$ by $n$ matrix. The rows of $A$ define \textit{hyperplanes} in $\R^n$ and the binary sign information tells us which side of the hyperplane each data point lies on. Throughout, we will take $A$ to have independent identically distributed standard Gaussian entries. 
Given $Q$ and $b$,
 we wish to train an algorithm that can be used to classify new signals, available only in a similar binary form via the matrix $A$, for which the label is unknown. 
 
 \subsection{Contribution} 
Our contribution is a \textit{framework} for classifying data into a given number of classes using only a binary representation of the data. This framework serves several purposes: (i) it provides mathematical tools that can be used for classification in applications where data is already captured in a simple binary representation, (ii) demonstrates that for general problems, classification can be done effectively using 
low-dimensional measurements, (iii) suggests an approach to use these measurements for classification using 
low computation, (iv) provides a simple technique for classification that can be mathematically anlayzed.  We believe this framework can be extended and utilized to build novel algorithmic approaches for many types of learning problems. In this work, we present one method for classification using training data, illustrate its promise on synthetic and real data, and provide a theoretical analysis of the proposed approach in the simple setting of two-dimensional signals and two possible classes. Under mild assumptions, we derive an explicit lower bound on the probability that a new data point gets classified correctly. This analysis serves as a foundation for analyzing the method in more complicated settings, and a framework for studying similar types of approaches.

\subsection{Organization} We proceed next in Section \ref{sec:prior} with a brief overview of related work. Then, in Section \ref{section::algorithm} we propose a two-stage method for classifying data into a given number of classes using only a binary representation of the data. The first stage of the method performs training on data with known class membership, and the second stage is used for classifying new data points with a priori unknown class membership. Next, in Section \ref{section::experiments} we demonstrate the potential of the proposed approach on both synthetically generated data as well as real datasets with application to handwritten digit recognition and facial recognition. Finally, in Section \ref{section::theory} we provide a theoretical analysis of the proposed approach in the simple setting of two-dimensional signals and two classes. We conclude in Section \ref{sec::conclude} with some discussion and future directions.

\subsection{Prior Work}\label{sec:prior}

There is a large body of work on several areas related to the subject of this paper, ranging from classification to compressed sensing, hashing, quantization, and deep learning. Due to the popularity and impact of each of these research areas, any review of prior work that we provide here must necessarily be non-exhaustive. Thus, in what follows, we briefly discuss related prior work, highlighting connections to our work but also stressing the distinctions.

Support vector machines (SVM) (see, e.g., \cite{CristS_Introduction, hearst1998support, andrew2000introduction, joachims1998text, steinwart2008support}) have become popular in machine learning, and are often used for classification. Provided a training set of data points and known labels, the SVM problem is to construct the optimal hyperplane (or hyperplanes) separating the data (if the data is linearly separable) or maximizing the geometric margin between the classes (if the data is not linearly separable). Although related, the approach taken in this paper is fundamentally different than in SVM. Instead of searching for the \textit{optimal} separating hyperplane, our proposed algorithm uses many, randomly selected hyperplanes (via the rows of the matrix $A$), and uses the relationship between these hyperplanes and the training data to construct a classification procedure that operates on information between the same hyperplanes and the data to be classified. 

The process of transforming high-dimensional data points into low-dimensional spaces has been studied extensively in related contexts. For example, the pioneering Johnson-Lindenstrauss Lemma states that any set of $p$ points in high dimensional Euclidean space can be (linearly) embedded into $O(\epsilon^{-2} \log(p))$ dimensions, without distorting the distance between any two points by more than a small factor, namely $\epsilon$  \cite{JohnsL_Extensions}. Since the original work of Johnson and Lindenstrauss, much work on Johnson-Lindenstrauss embeddings (often motivated by signal processing and data analysis applications) has focused on randomized embeddings where the matrix associated with the linear embedding is drawn from an appropriate random distribution. Such random embeddings include those based on Gaussian and other subgaussian random variables as well as those that admit fast implementations, usually based on the fast Fourier transform (see, e.g., \cite{ailon2006approximate, achlioptas2003database, dasgupta2003elementary}). 

Another important line of related work is \textit{compressed sensing}, in which it has been demonstrated that far fewer linear measurements than dictated by traditional Nyquist sampling can be used to represent high-dimensional data  \cite{CandeRT_Stable,CandeRT_Robust,Donoh_Compressed}. For a signal $x\in\R^n$, one obtains $m<n$ measurements of the form $y = Ax$ (or noisy measurements $y=Ax+z$ for $z\in\R^m$), where $A\in\R^{m\times n}$, and the goal is to recover the signal $x$. By assuming the signal $x$ is $s$-sparse, meaning that $\| x\|_0 = |\supp(x)| = s \ll n$, the recovery problem becomes well-posed under certain conditions on $A$. Indeed, there is now a vast literature describing recovery results and algorithms when $A$, say, is a random matrix drawn from appropriate distributions (including those where the entries of $A$ are independent Gaussian random variables). The relationship between Johnson-Lindenstrauss embeddings and compressed sensing is deep and bi-directional; matrices that yield Johnson-Lindenstrauss embeddings make excellent compressed sensing matrices \cite{baraniuk2006johnson} and conversely, compressed sensing matrices (with minor modifications) yield Johnson-Lindenstrauss embeddings \cite{krahmer2011new}.

To allow processing on digital computers, compressive measurements must often be \textit{quantized}, or mapped to discrete values from some finite set. The extreme quantization setting where only the sign bit is acquired is known as \textit{one-bit compressed sensing} and was introduced in \cite{BoufoB_1Bit}. In this framework, the measurements now take the form $y = \sign(Ax)$, and the objective is still to recover the signal $x$. Several methods have since been developed to recover the signal $x$ (up to normalization) from such simple one-bit measurements (see e.g., \cite{PlanV_One,PlanV_Robust,gopi2013one,JacquLBB_Robust,yan2012robust,jacques2013quantized}). Although the data we consider in this paper takes a similar form, the overall goal is different; rather than signal \textit{reconstruction}, our interest is data \textit{classification}.

More recently, there has been growing interest in binary embeddings (embeddings into the binary cube, see e.g., \cite{PlanV_Dimension,yu2014circulant,gong2013iterative,price2015binary,choromanska2016binary, dirksen2016fast}), where it has been observed that using certain linear projections and then applying the sign operator as a nonlinear map largely preserves information about the angular distance between vectors provided one takes sufficiently many measurements. Indeed, the measurement operators used for binary embeddings are Johnson-Lindenstrauss embeddings and thus also similar to those used in compressed sensing, so they again range from random Gaussian and subgaussian matrices to those admitting fast linear transformations, such as random circulant matrices (see, e.g., \cite{dirksen2016fast} for an overview). Although we consider a similar binary measurement process, we are not necessarily concerned with geometry preservation in the low-dimensional space, but rather the ability to still perform data classification.

Deep Learning is an area of machine learning based on learning data representations using multiple levels of abstraction, or layers. Each of these layers is essentially a function whose parameters are learned, and the full network is thus a composition of such functions.  Algorithms for such deep neural networks  have recently obtained state of the art results for classification. Their success has been due to the availability of large training data sets coupled with advancements in computing power and the development of new techniques (e.g., \cite{krizhevsky2012imagenet,simonyan2014very,szegedy2015going,russakovsky2015imagenet}). We consider deep learning and neural networks as motivational to our layered algorithm design.  However, we are not tuning nor optimizing parameters as is typically done in deep learning, nor do our layers necessarily possess the structure typical in deep learning ``architectures"; this makes our approach potentially simpler and easier to work with.

\section{The Proposed Classification Algorithm} \label{section::algorithm}
The training phase of our algorithm is detailed in Algorithm \ref{proposed algorithm1}, where we suppose the training data $Q = \sign(AX)$ and associated labels $b$ are available. Indeed, the training algorithm proceeds in $L$ ``layers". 
In the $\ell$-th layer, $m$ index sets $\GamLi{\ell}{i} \subset [m]$,  $|\GamLi{\ell}{i}| = \ell$, $i=1,...,m$, are randomly selected, 
so that all elements of $\GamLi{\ell}{i}$ are unique, and $\GamLi{\ell}{i} \neq \GamLi{\ell}{j}$ for $i\neq j$.
This is achieved by selecting the multi-set of $\Lambda_{\ell,i}$'s uniformly at random from a set of cardinality ${{m}\choose{\ell}}\choose m$.
During the $i$-th ``iteration" of the $\ell$-th layer, the rows of $Q$ indexed by $\GamLi{\ell}{i}$ are used to form the $\ell \times p$ matrix $Q^{\GamLi{\ell}{i}} \in \{\pm 1\}^{\ell \times p}$, and the unique sign patterns $q \in \{\pm 1\}^\ell$ are extracted from the columns of $Q^{\GamLi{\ell}{i}}$. The number of unique sign patterns (i.e., distinct columns) in $Q^{\GamLi{\ell}{i}}$ is given by $T_{\ell,i}\in \mathbb{N}$.

For example, 
at the first layer the possible unique sign patterns are 1 and -1, describing which side of the selected hyperplane the training data points lie on; at the second layer the possible unique sign patters are $\begin{bmatrix} 1 \\ 1 \end{bmatrix}$, $\begin{bmatrix} 1 \\ -1 \end{bmatrix}$, $\begin{bmatrix} -1 \\ 1 \end{bmatrix}$, $\begin{bmatrix} -1 \\ -1 \end{bmatrix}$, describing which side of the two selected hyperplanes the training data points lie on, and so on for the subsequent layers. For the $t$-th sign pattern and $g$-th class, a \textit{\RF} parameter $r(\ell,i,t,g)$ that uses knowledge of the number of training points in class $g$ having the $t$-th sign pattern, is calculated for every $\GamLi{\ell}{i}$. Larger values of  $r(\ell,i,t,g)$ suggest that the $t$-th sign pattern is more heavily dominated by class $g$; thus, if a signal with unknown label corresponds to the $t$-th sign pattern, we will be more likely to classify it into the $g$-th class. In this paper, we use the following choice for the \RF parameter $r(\ell,i,t,g)$, which we found to work well experimentally. Below, $P_{g|t}$ denotes the number of training points from the $g$-th class with the $t$-th sign pattern at the $i$-th set selection in the $\ell$-th layer (i.e., the $t$-th sign pattern determined from the set selection $\GamLi{\ell}{i}$): 
\begin{align} \label{RF3}
r(\ell,i,t,g) &= \frac{P_{g|t}}{\sum_{j=1}^G P_{j|t}} \frac{\sum_{j=1}^G |P_{g|t} - P_{j|t}|}{\sum_{j=1}^G P_{j|t}}.
\end{align}
Let us briefly explain the intuition for this formula. The first fraction in \eqref{RF3} indicates the proportion of training points in class $g$ out of all points with sign pattern $t$. The second fraction in \eqref{RF3} is a balancing term that gives more weight to group $g$ when that group is much different in size than the others with the same sign pattern. 
If $P_{j|t}$ is the same for all classes $j = 1,\dots,G$, then $r(\ell,i,t,g)=0$ for all $g$, and thus no class is given extra weight for the given sign pattern, set selection, and layer. 
If $P_{g|t}$ is nonzero and $P_{j|t} = 0$ for all other classes, then $r(\ell,i,t,g) = G-1$ 
  and $r(\ell,i,t,j) = 0$ for all $j\neq g$, so that class $g$ receives the largest weight.

\begin{algorithm}[ht]
\caption{Training 
} 
\label{proposed algorithm1}
\begin{algorithmic}
\STATE \textbf{input:} binary training data $Q$, training labels $b$, number of classes $G$, number of layers $L$
\STATE 
\FOR{$\ell$ from 1 to $L$, $i$ from 1 to $m$}
\STATE
\begin{tabular}{ll}
\textbf{select:} & Randomly select $\GamLi{\ell}{i} \subset [m]$, $|\GamLi{\ell}{i}| = \ell$ \\
\textbf{determine:} & Determine the $T_{\ell,i}\in\mathbb{N}$ unique column patterns in $Q^{\GamLi{\ell}{i}}$

\end{tabular}
\FOR{$t$ from 1 to $T_{\ell,i}$, $g$ from 1 to $G$}
\STATE
\begin{tabular}{ll}
\textbf{compute:} & Compute $r(\ell,i,t,g)$ by (\ref{RF3})\\
\end{tabular}
\ENDFOR
\ENDFOR
\end{algorithmic}
\end{algorithm}

Once the algorithm has been trained, we can use it to classify new signals. Suppose $x\in\R^n$ is a new signal for which the class is unknown, and we have available the quantized measurements $q = \sign(Ax)$. Then Algorithm \ref{proposed algorithm2} is used for the classification of $x$ into one of the $G$ classes. Notice that the number of layers $L$, the learned \RF values $r(\ell,i,t,g)$, the number of unique sign patterns $T_{\ell,i}$, and the set selections $\GamLi{\ell}{i}$ at each iteration of each layer are all available from Algorithm \ref{proposed algorithm1}. First, the decision vector $\tilde{r}$ is initialized to the zero vector in $\R^G$.  Then for each layer $\ell$ and set selection $i$, 
the sign pattern $q^{\GamLi{\ell}{i}}$ is determined and the index $t^\star\in[T_{\ell,i}]$ is identified corresponding to the sign patterns that were determined during training.  For each class $g$,  $\tilde{r}(g)$ is updated via $\tilde{r}(g) \leftarrow \tilde{r}(g) + r(\ell,i,t^\star,g)$. If it happens that the sign pattern for $x$ does not match any sign pattern determined during training, no update to $\tilde{r}$ is performed. Finally, after scaling $\tilde{r}$ with respect to the number of layers and measurements, the largest entry of $\tilde{r}$ identifies how the estimated label $\widehat{b}_x$ of $x$ is set. Note that this scaling does not actually affect the outcome of classification, we use it simply to ensure the quantity does not become unbounded for large problem sizes.

\begin{algorithm}[ht]
\caption{Classification} 
\label{proposed algorithm2}
\begin{algorithmic}
\STATE \textbf{input:} binary data $q$, number of classes $G$, number of layers $L$, learned parameters $r(\ell,i,t,g)$, $T_{\ell,i}$, and $\GamLi{\ell}{i}$ from Algorithm \ref{proposed algorithm1}
\STATE 
\STATE \textbf{initialize:} $\tilde{r}(g) = 0$ for $g = 1,\dots,G$.
\FOR{$\ell$ from 1 to $L$, $i$ from 1 to $m$}
\STATE
\begin{tabular}{ll}
\textbf{identify:} & Identify the pattern $t^\star\in[T_{\ell,i}]$ to which $q^{\GamLi{\ell}{i}}$ corresponds \\
\end{tabular}
\FOR{$g$ from 1 to $G$} 
\STATE
\begin{tabular}{ll}
\textbf{update:} & $\tilde{r}(g) = \tilde{r}(g) + r(\ell,i,t^\star,g)$ \\
\end{tabular}
\ENDFOR
\ENDFOR
\STATE \textbf{scale:} Set $\tilde{r}(g) = \frac{\tilde{r}(g)}{Lm}$ for $g=1,\dots,G$ 
\STATE \textbf{classify:} $\widehat{b}_x = \argmax_{g\in\{1,\dots,G\}}\tilde{r}(g)$
\end{algorithmic}
\end{algorithm}

\section{Experimental Results}\label{section::experiments}
In this section, we provide experimental results of Algorithms \ref{proposed algorithm1} and \ref{proposed algorithm2} for synthetically generated datasets, handwritten digit recognition using the MNIST dataset, and facial recognition using the extended YaleB database. 

In all of the experiments, the matrix $A$ is taken to have i.i.d. standard Gaussian entries. Also, we assume the data is centered. To ensure this, a pre-processing step on the raw data is performed to account for the fact that the data may not be centered around the origin. That is, given the original training data matrix $X$, we calculate $\mu = \frac{1}{p} \sum_{i=1}^p x_i$. Then for each column $x_i$ of $X$, we set $x_i \leftarrow x_i - \mu$. The testing data is adjusted similarly by $\mu$. Note that this assumption can be overcome in future work by using \textit{dithers}---that is, hyperplane dither values may be learned so that $Q = \sign(AX + \tau)$, where $\tau\in\R^m$---or by allowing for pre-processing of the data.

\subsection{Classification of Synthetic Datasets}
\begin{figure}[!htbp]
\centering
\begin{tabular}{cc}
\includegraphics[height=2in]{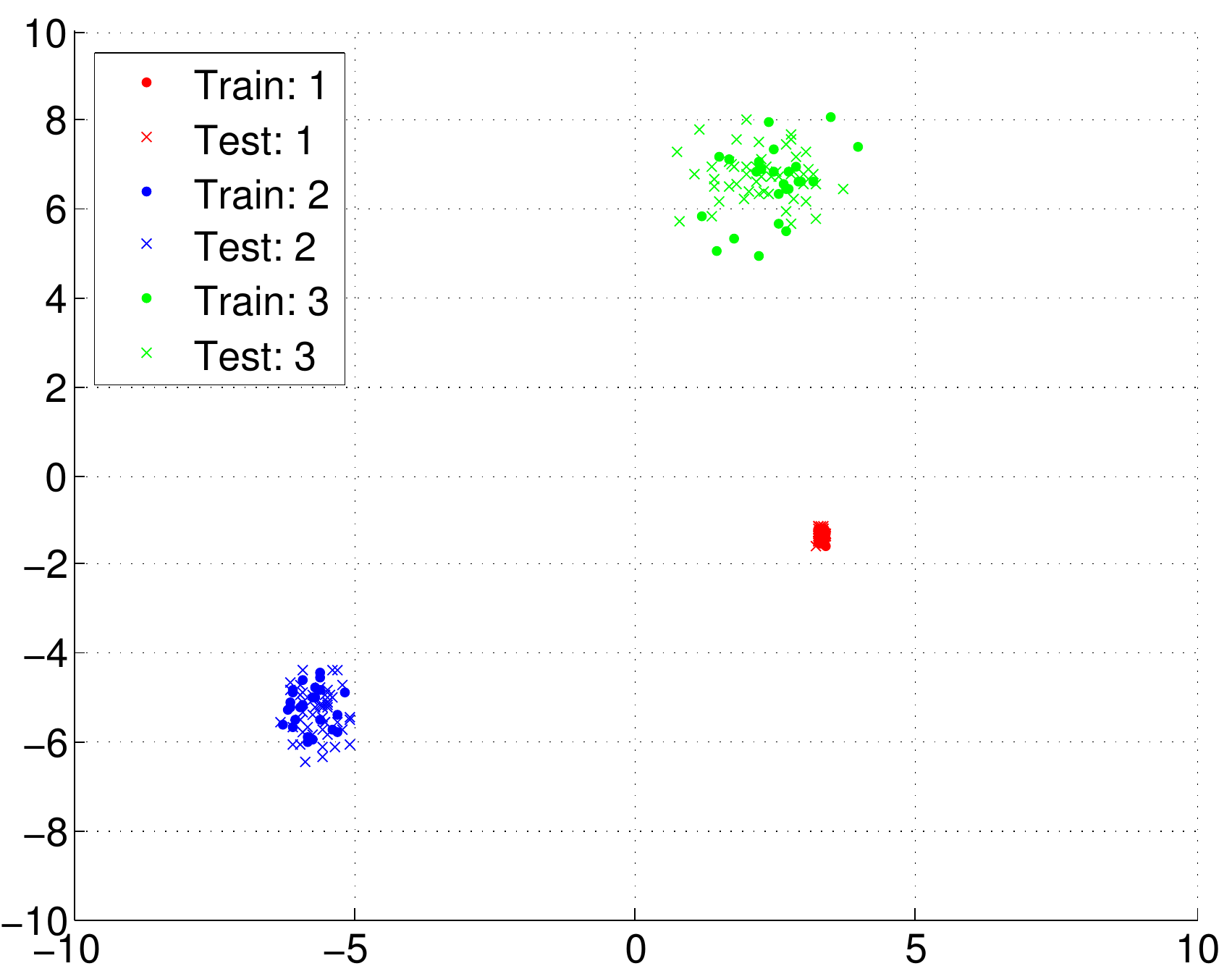} &
\includegraphics[height=2in]{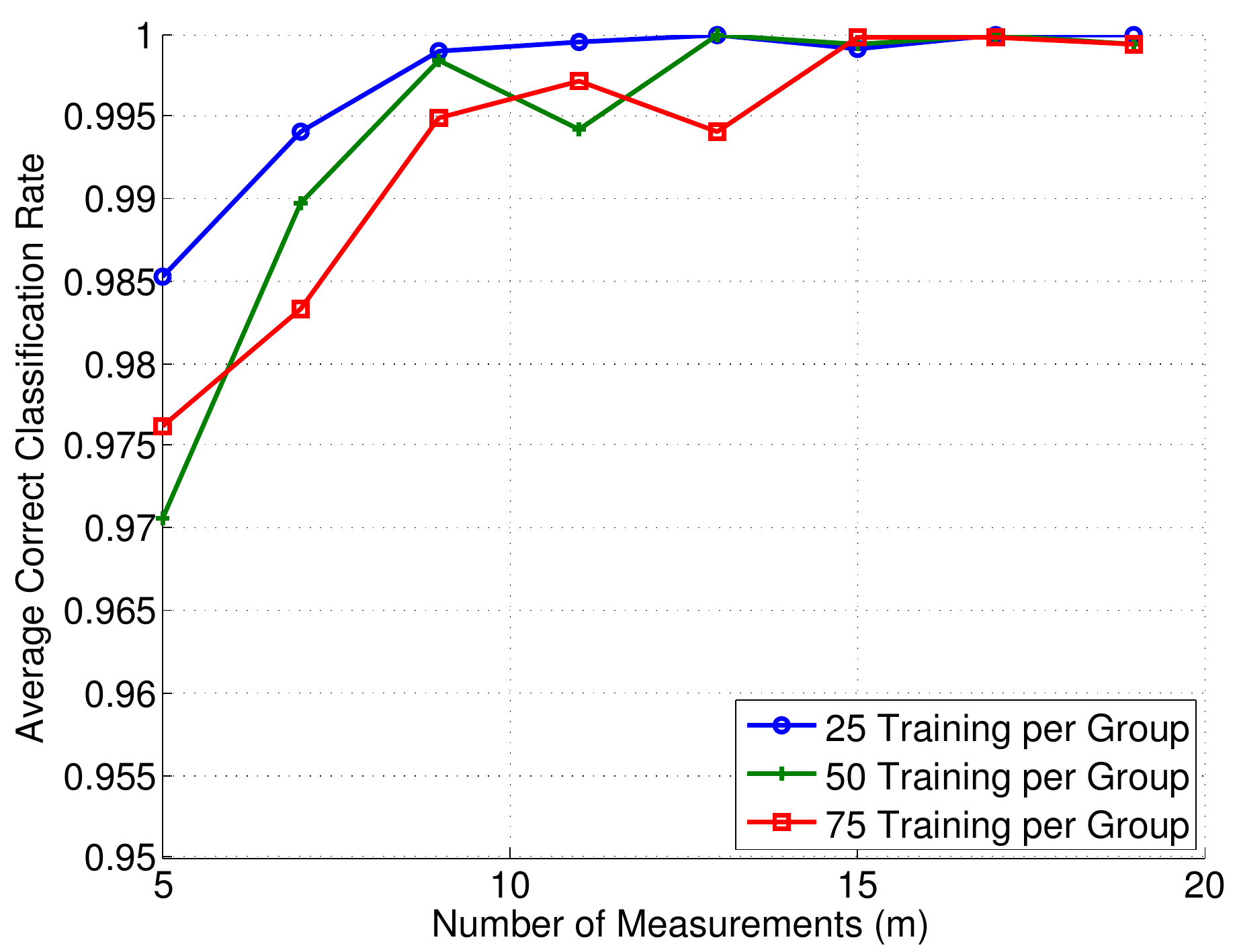} \\
\end{tabular}
\caption{Synthetic classification experiment with three Gaussian clouds ($G=3$), $L=1$, $n=2$, 50 test points per group, and 30 trials of randomly generating $A$. (Left) Example training and testing data setup. (Right) Average correct classification rate versus $m$ and for the indicated number of training points per class.}
\label{syn:gaussian clouds}
\end{figure}

In our first stylized experiment, we consider three classes of Gaussian clouds in $\R^2$ (i.e., $n=2$); see Figure \ref{syn:gaussian clouds} for an example training and testing data setup. For each choice of $m\in\{5,7,9,11,13,15,17,19\}$ and $p\in\{75,150,225\}$ with equally sized training data sets for each class (that is, each class is tested with either 25, 50, or 75 training points), 
 we execute Algorithms \ref{proposed algorithm1} and \ref{proposed algorithm2} with a single layer and 30 trials of generating $A$. We perform classification of 50 test points per group, and report the average correct classification rate over all trials. The right plot of Figure \ref{syn:gaussian clouds} shows that $m\geq 15$ results in nearly perfect classification. 

\begin{figure}[!htbp]
\centering
\begin{tabular}{cc}
\includegraphics[height=2in]{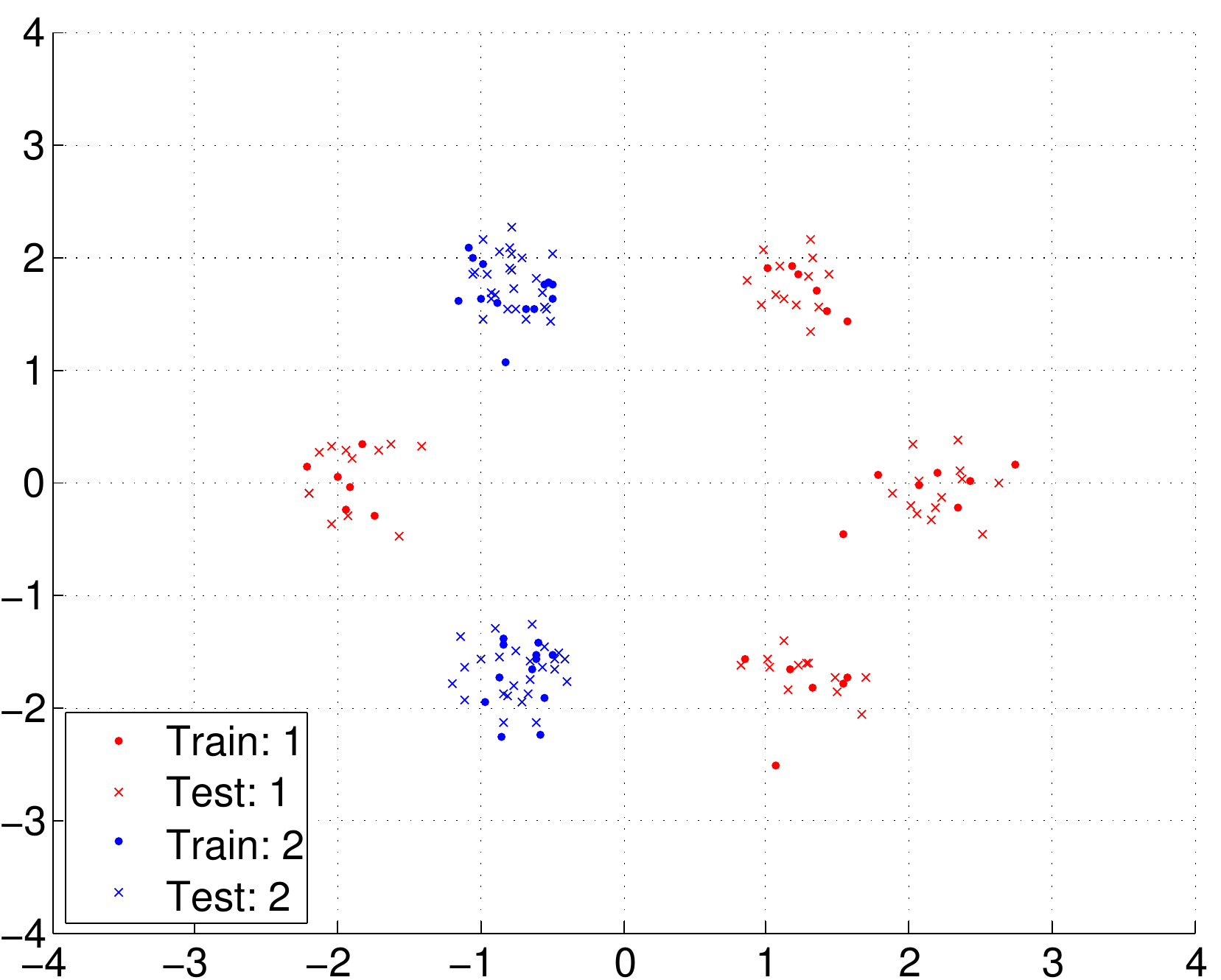} &
\includegraphics[height=2in]{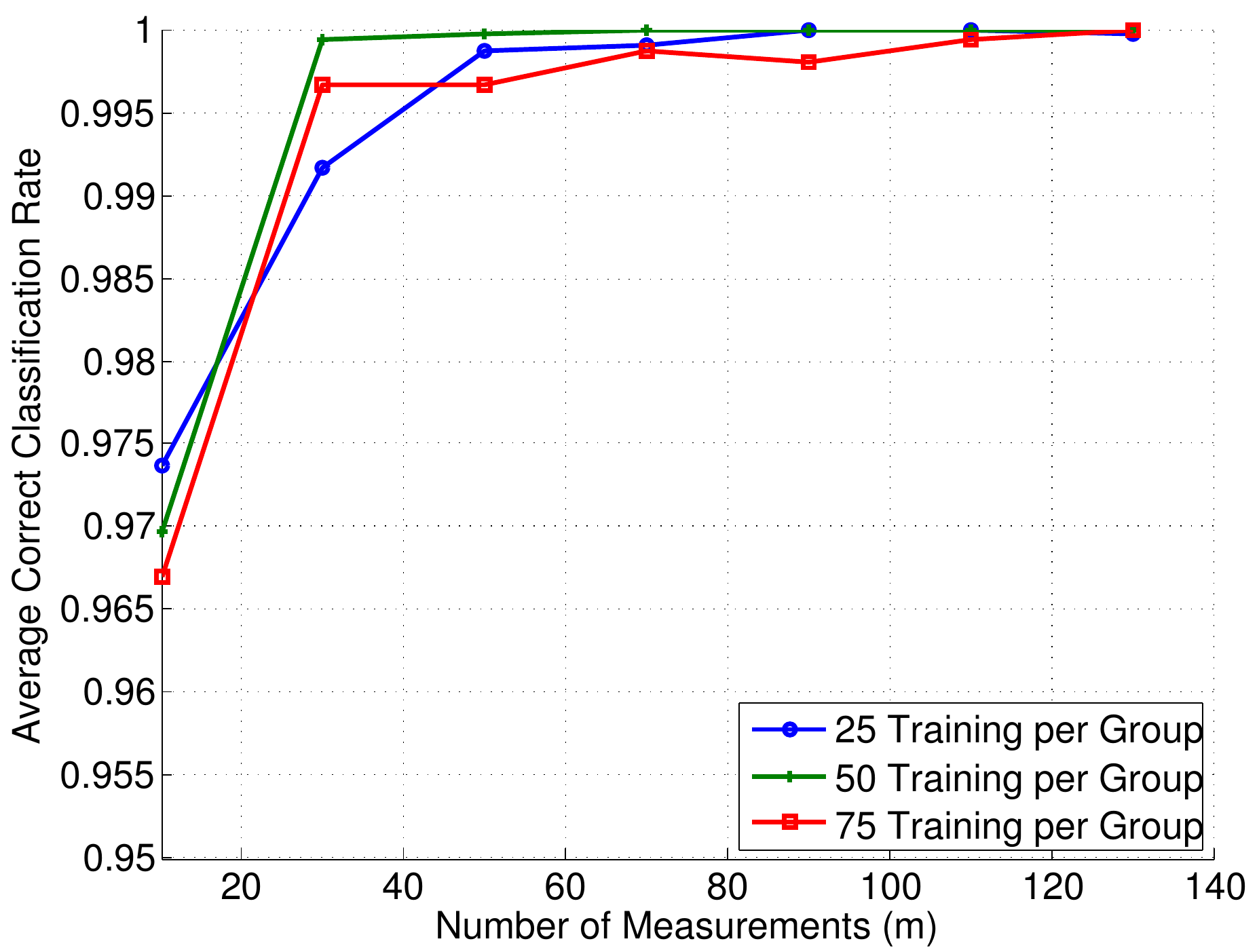} \\
\end{tabular}
\caption{Synthetic classification experiment with six Gaussian clouds and two classes ($G=2$), $L=4$, $n=2$, 50 test points per group, and 30 trials of randomly generating $A$. (Left) Example training and testing data setup. (Right) Average correct classification rate versus $m$ and for the indicated number of training points per class.}
\label{syn:gaussian2 alternating6}
\end{figure}

\begin{figure}[!htbp]
\centering
\begin{tabular}{cc}
\includegraphics[height=2in]{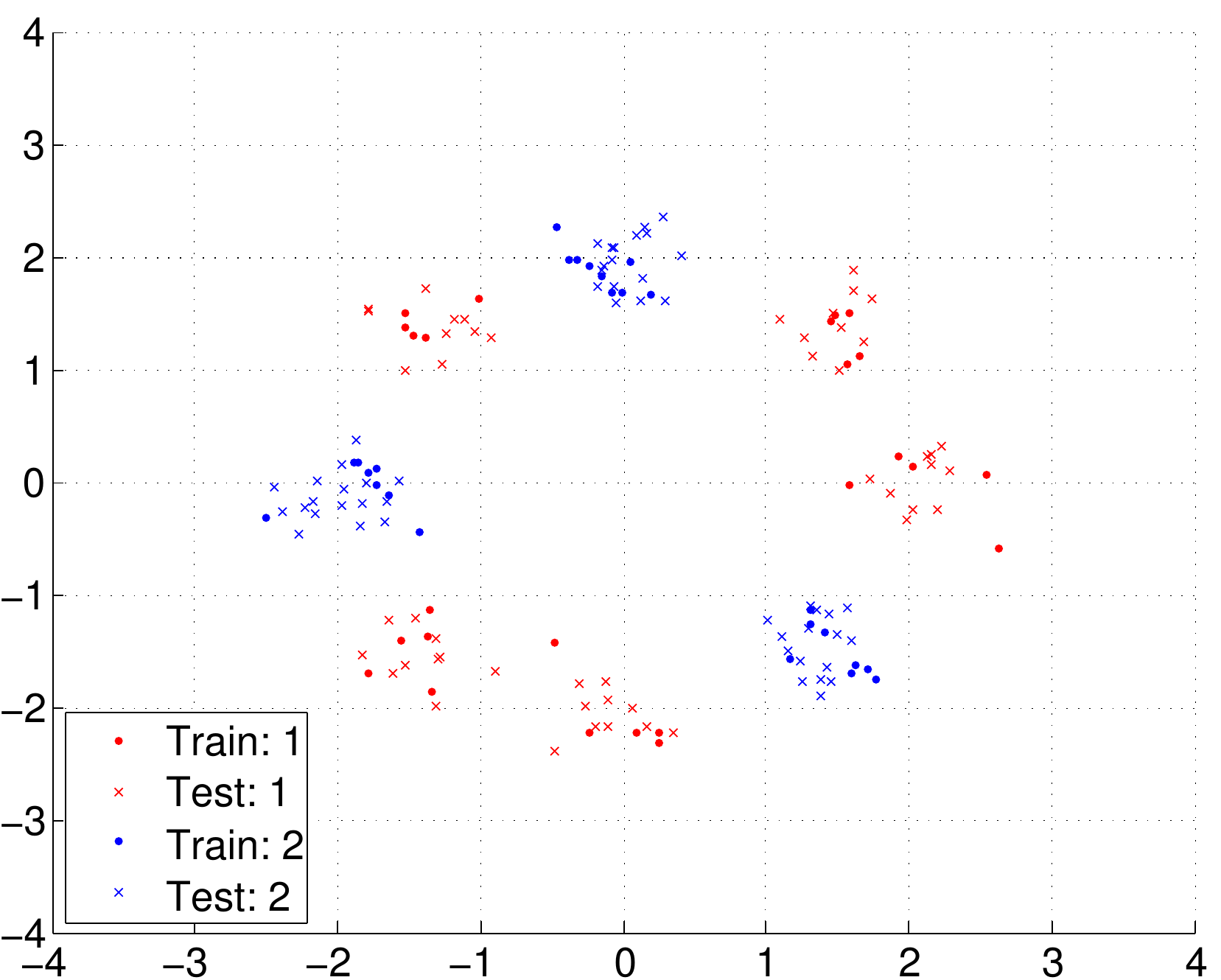} &
\includegraphics[height=2in]{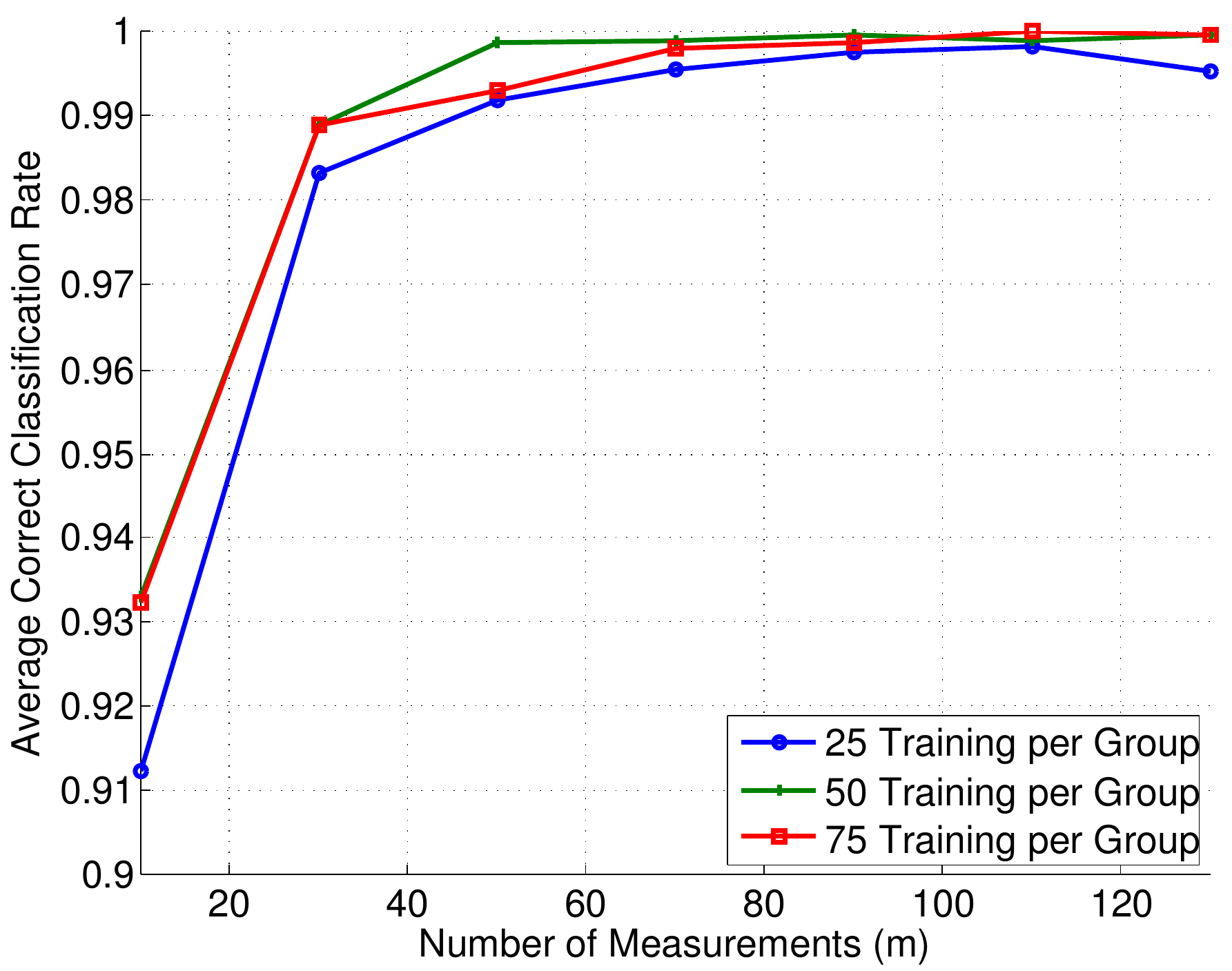} \\
\end{tabular}
\caption{Synthetic classification experiment with eight Gaussian clouds and two classes ($G=2$), $L=5$, $n=2$, 50 test points per group, and 30 trials of randomly generating $A$. (Left) Example training and testing data setup. (Right) Average correct classification rate versus $m$ and for the indicated number of training points per class.}
\label{syn:gaussian2 alternating8}
\end{figure}

Next, we present a suite of experiments where we again construct the classes as Gaussian clouds in $\R^2$, but utilize a non-uniformly alternating pattern around the origin with respect to the classes. In each case,  we set the number of training data points for each class to be 25, 50, and 75. In Figure \ref{syn:gaussian2 alternating6}, we have two classes forming a total of six Gaussian clouds, and execute Algorithms \ref{proposed algorithm1} and \ref{proposed algorithm2} using four layers and $m\in\{10,30,50,70,90,110,130\}$. The classification accuracy increases for larger $m$, with nearly perfect classification for the largest values of $m$ selected. 
 A similar experiment is shown in Figure \ref{syn:gaussian2 alternating8}, where we have two classes forming a total of eight Gaussian clouds, and execute the proposed algorithm using five layers.

\begin{figure}[!htbp]
\centering
\includegraphics[height=2in]{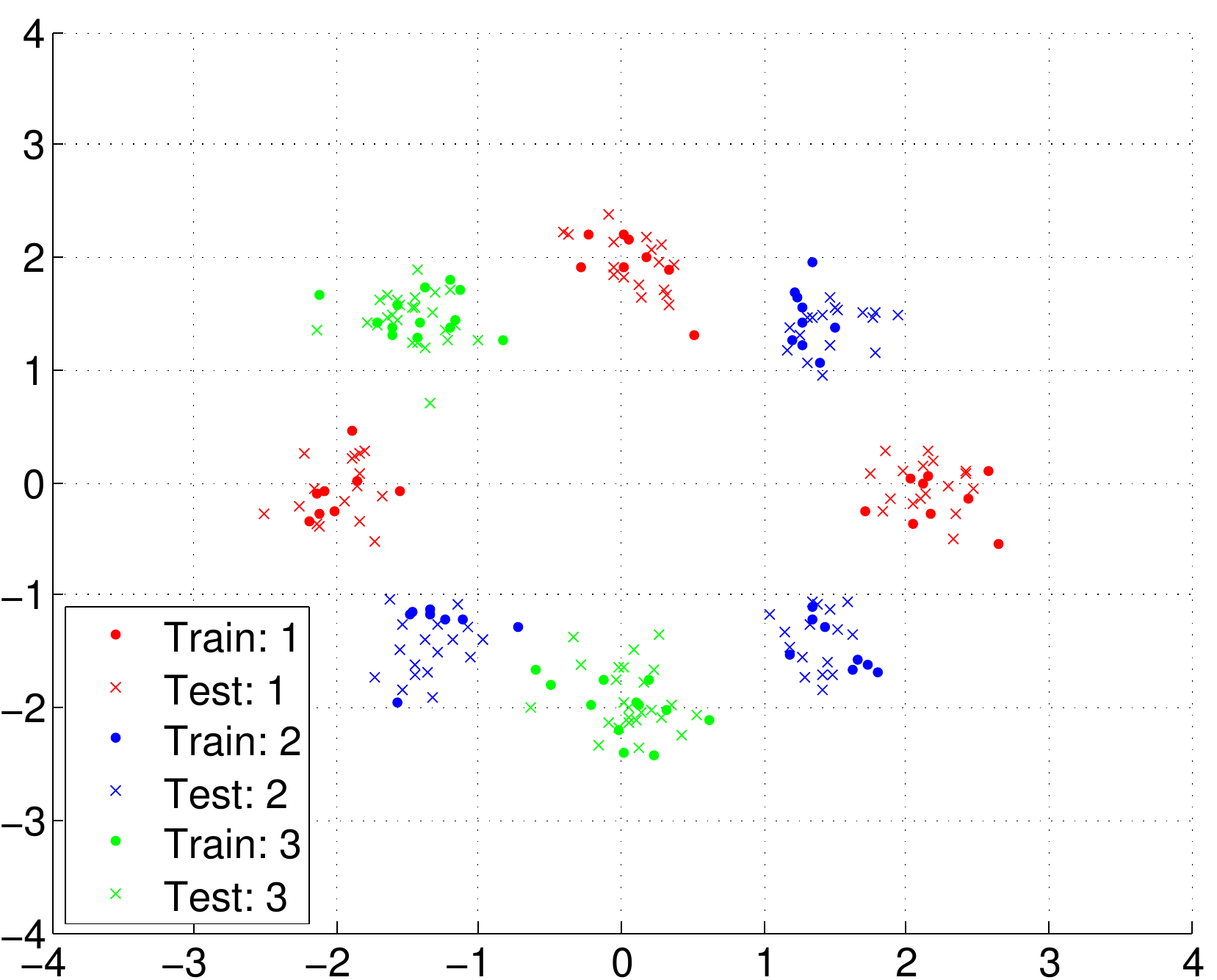} 
\begin{tabular}{cc}
\includegraphics[height=2in]{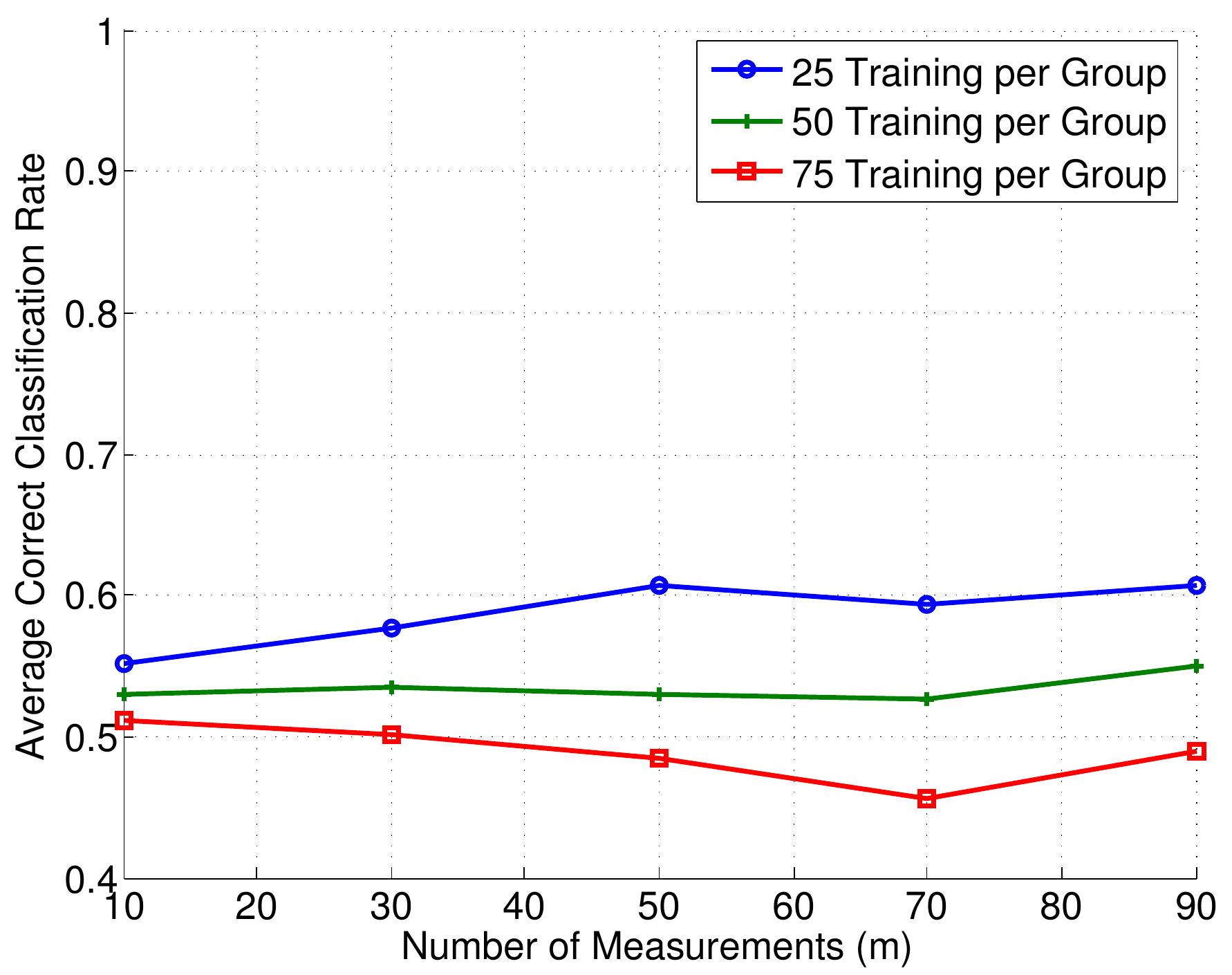} &
\includegraphics[height=2in]{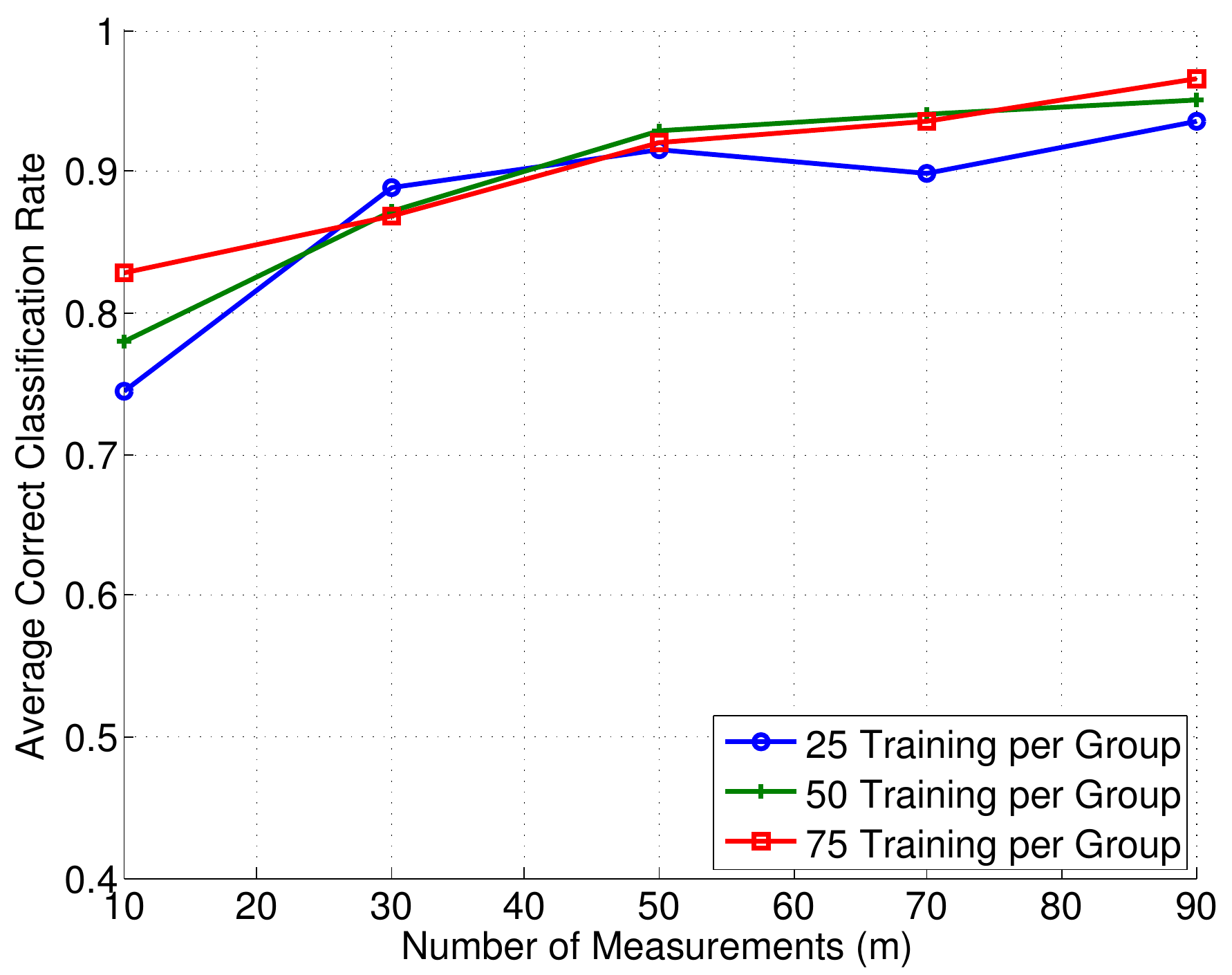} \\
(a) $L=1$ & (b) $L=2$ \\
\includegraphics[height=2in]{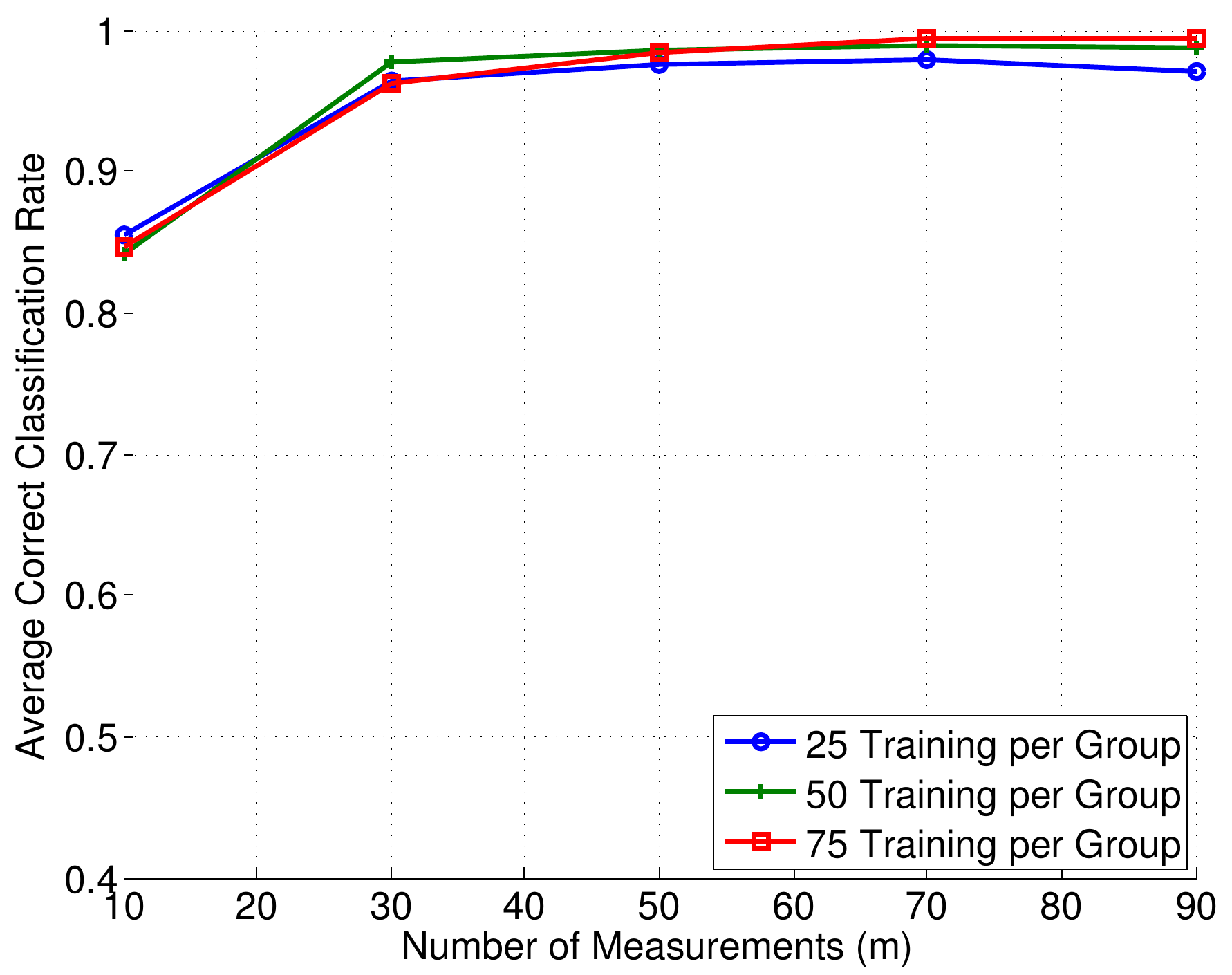} &
\includegraphics[height=2in]{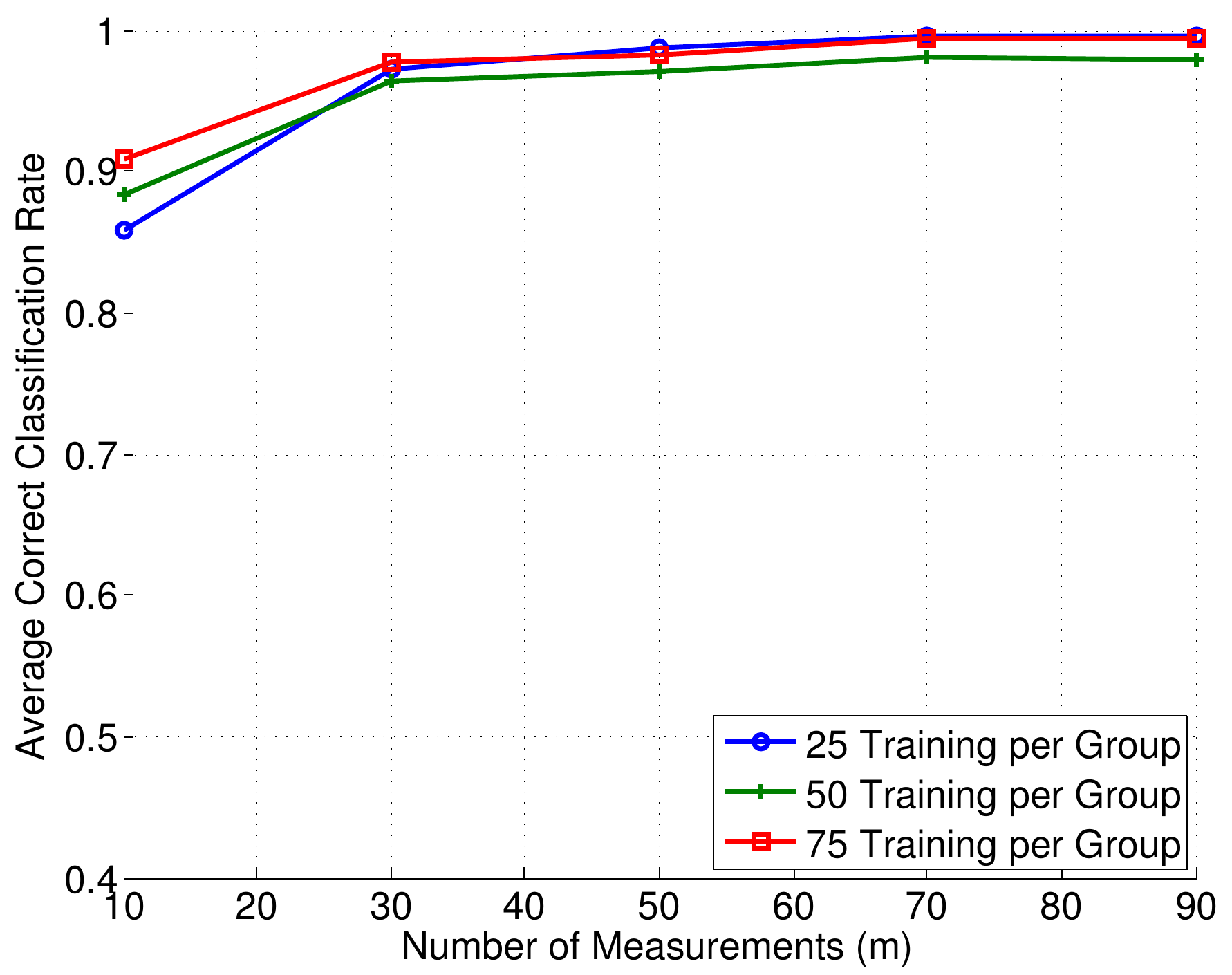} \\
(c) $L=3$ & (d) $L=4$ \\
\end{tabular}
\caption{Synthetic classification experiment with eight Gaussian clouds and three classes ($G=3$), $L=1,\dots,4$, $n=2$, 50 test points per group, and 30 trials of randomly generating $A$. (Top) Example training and testing data setup. Average correct classification rate versus $m$ and for the indicated number of training points per class for: (middle left) $L=1$, (middle right) $L=2$, (bottom left) $L=3$, (bottom right) $L=4$.}
\label{syn:gaussian3 alternating8}
\end{figure}

\begin{figure}[!htbp]
\centering
\includegraphics[height=2in]{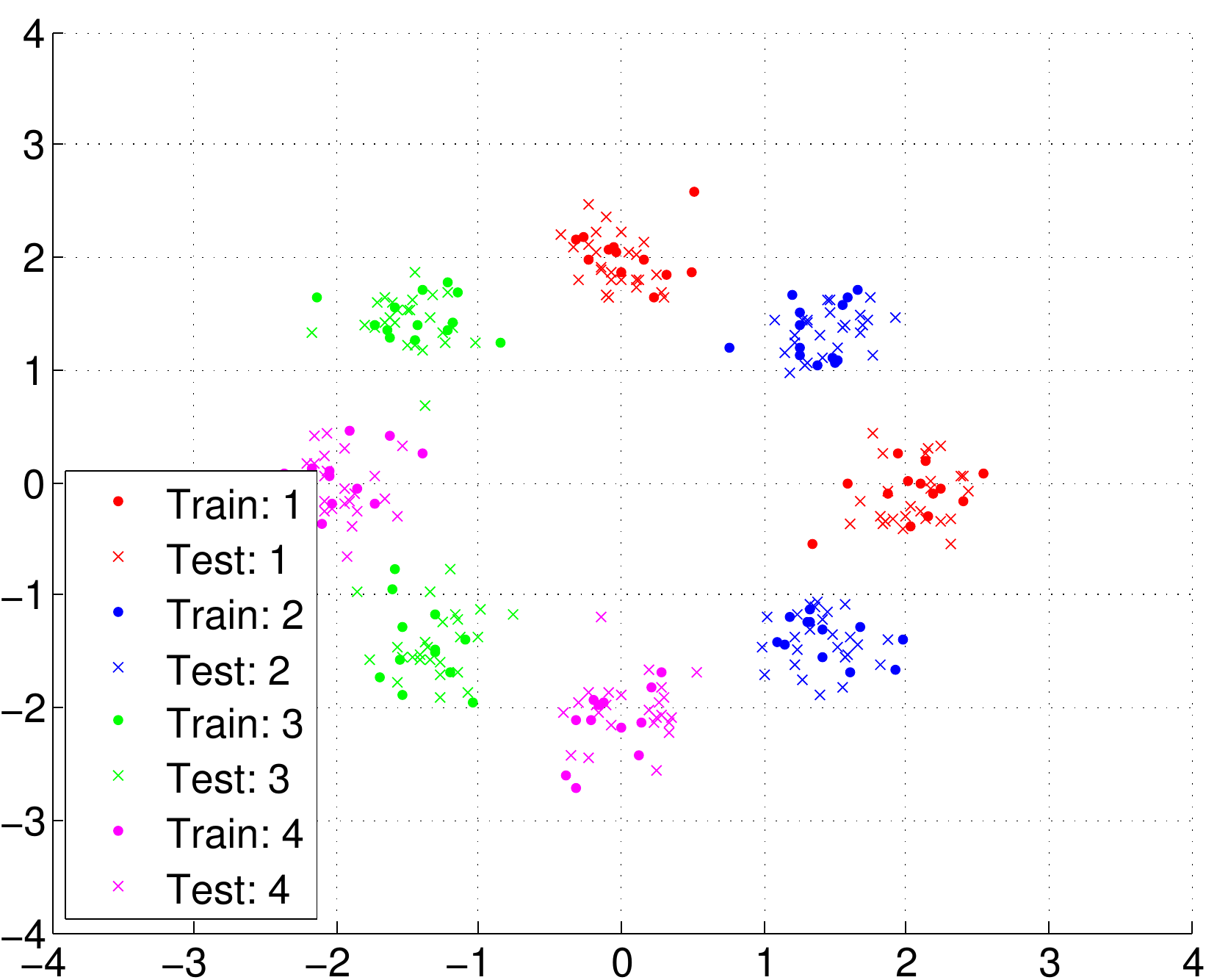} 
\begin{tabular}{cc}
\includegraphics[height=2in]{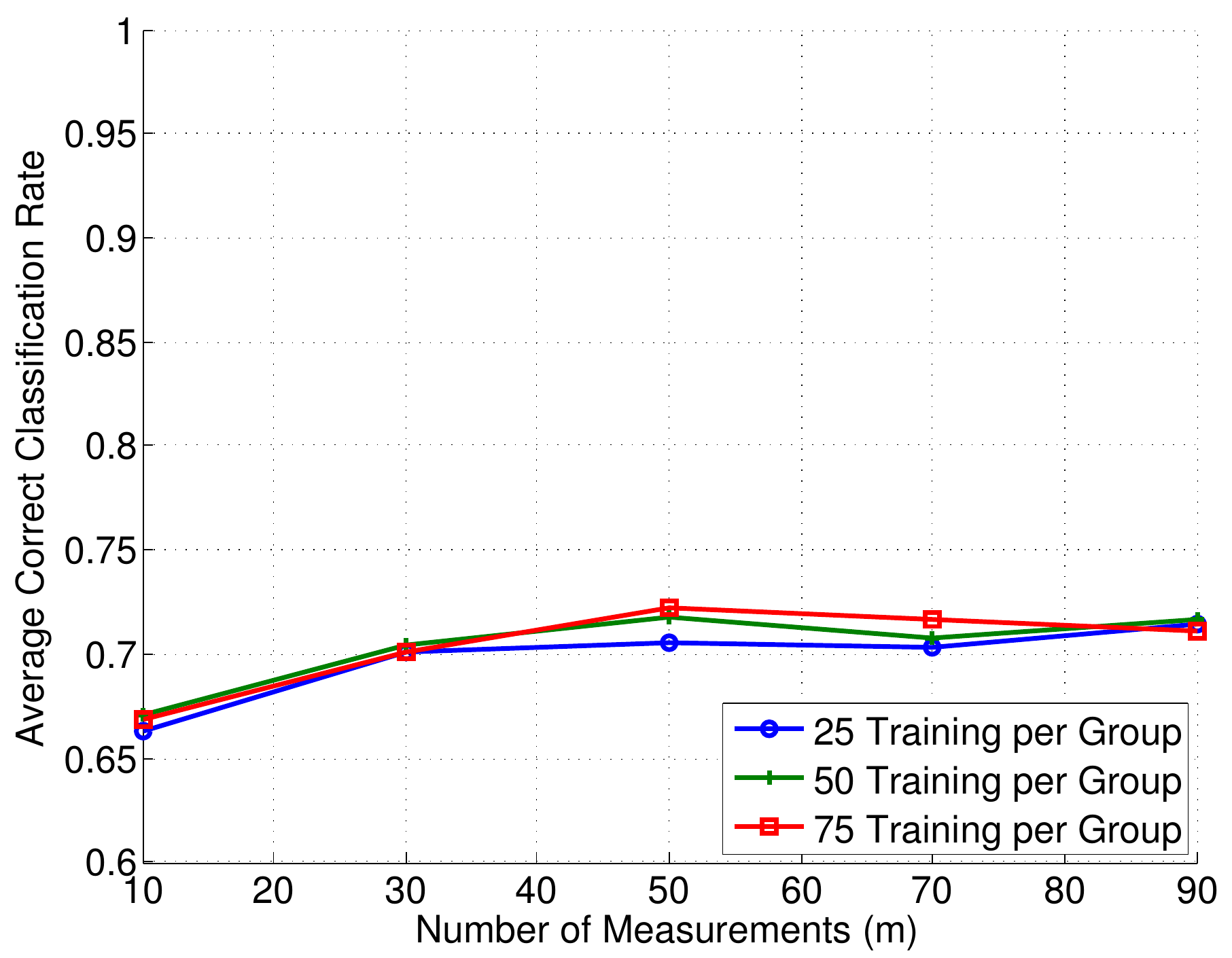} &
\includegraphics[height=2in]{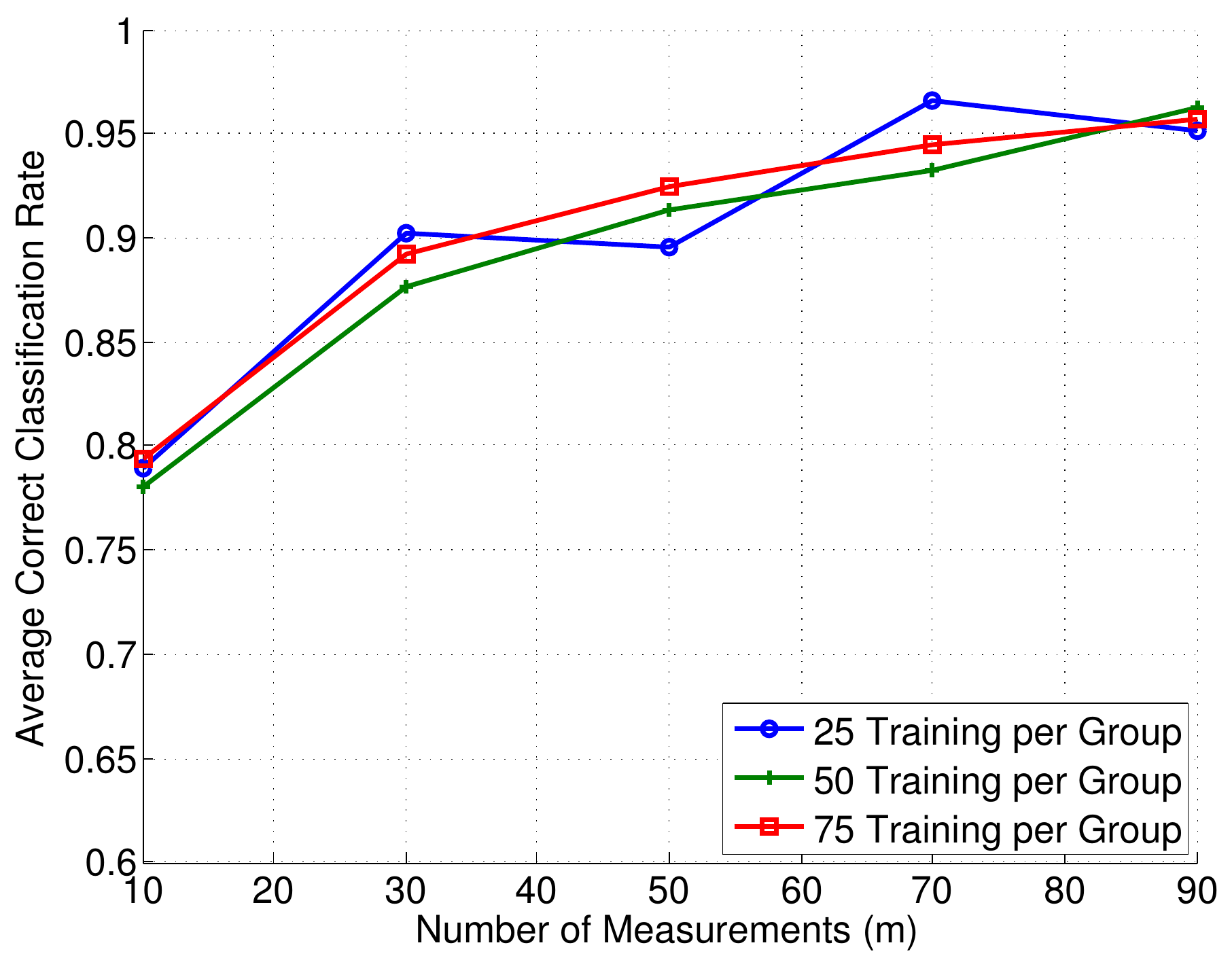} \\
(a) $L=1$ & (b) $L=2$ \\
\includegraphics[height=2in]{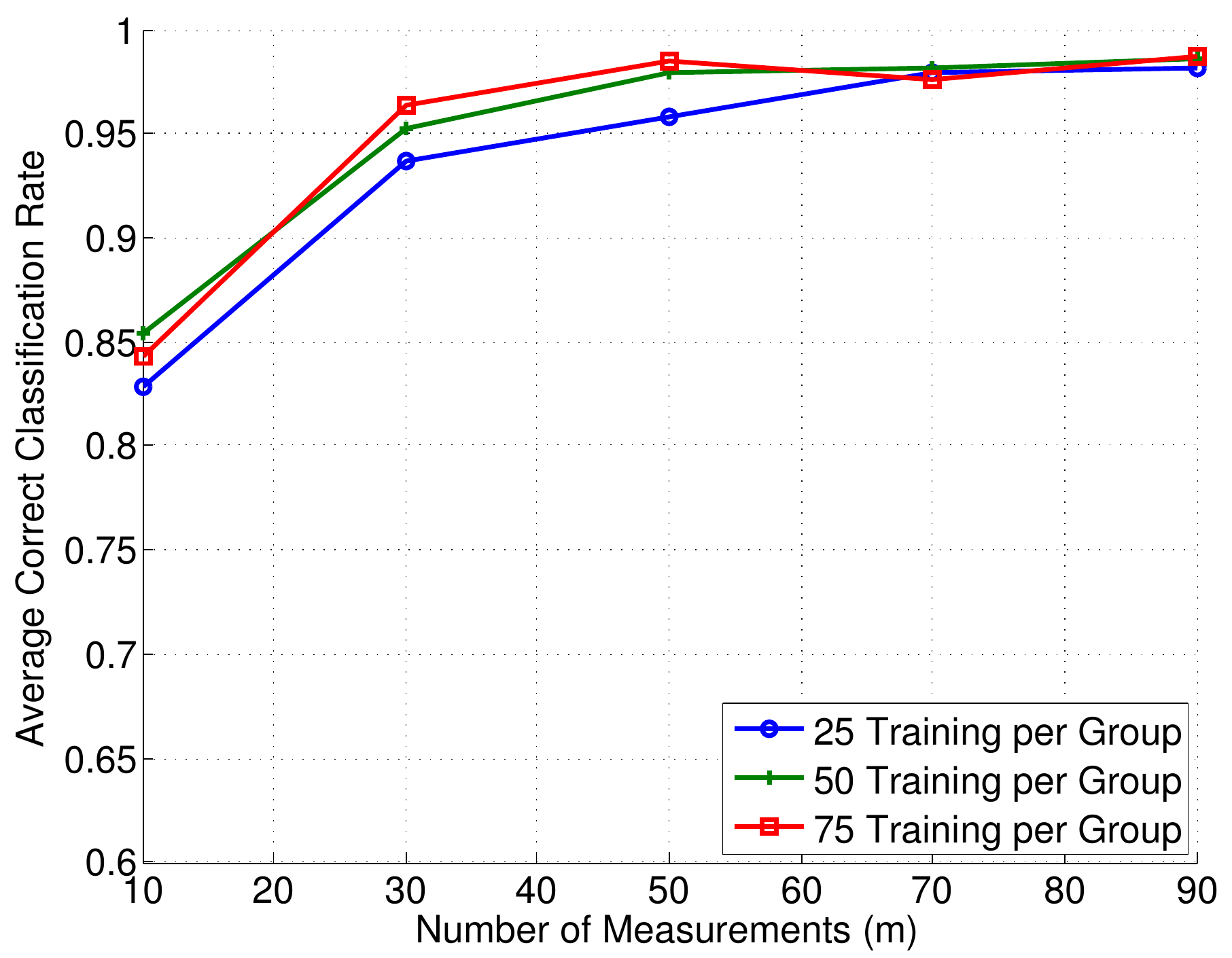} &
\includegraphics[height=2in]{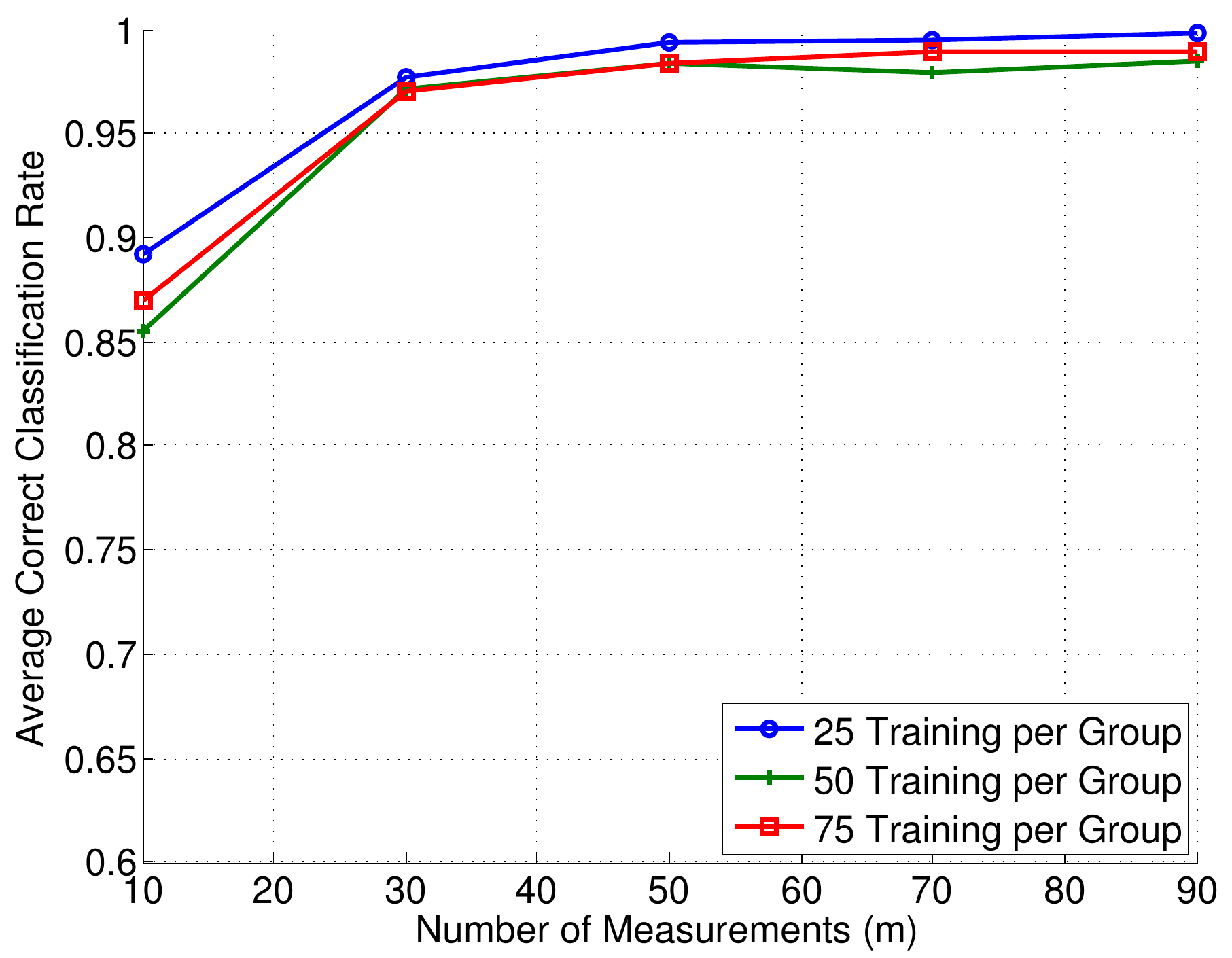} \\
(c) $L=3$ & (d) $L=4$ \\
\end{tabular}
\caption{Synthetic classification experiment with eight Gaussian clouds and four classes ($G=4$), $L=1,\dots,4$, $n=2$, 50 test points per group, and 30 trials of randomly generating $A$. (Top) Example training and testing data setup. Average correct classification rate versus $m$ and for the indicated number of training points per class for: (middle left) $L=1$, (middle right) $L=2$, (bottom left) $L=3$, (bottom right) $L=4$.}
\label{syn:gaussian4 alternating8}
\end{figure}

In the next two experiments, we display the classification results of Algorithms \ref{proposed algorithm1} and \ref{proposed algorithm2} when using $m\in\{10,30,50,70,90\}$ and one through four layers, and see that adding layers can be beneficial for more complicated data geometries.
In Figure \ref{syn:gaussian3 alternating8}, we have three classes forming a total of eight Gaussian clouds. We see that from both $L=1$ to $L=2$ and $L=2$ to $L=3$, there are huge gains in classification accuracy. In Figure \ref{syn:gaussian4 alternating8}, we have four classes forming a total of eight Gaussian clouds. Again, from both $L=1$ to $L=2$ and $L=2$ to $L=3$ we see large improvements in classification accuracy, yet still better classification with $L=4$. We note here that in this case it also appears that more training data does not improve the performance (and perhaps even slightly decreases accuracy); this is of course unexpected in practice, but we believe this happens here only because of the construction of the Gaussian clouds -- more training data leads to more outliers in each cloud, making the sets harder to separate.

\clearpage
\subsection{Handwritten Digit Classification}

In this section, we apply Algorithms \ref{proposed algorithm1} and \ref{proposed algorithm2} to the MNIST \cite{MNIST} dataset, which is a benchmark dataset of images of handwritten digits, each with $28 \times 28$ pixels. In total, the dataset has $60,000$ training examples and $10,000$ testing examples. 

\begin{figure}[!htbp]
\centering
\begin{tabular}{cc}
\raisebox{1.1\height}{\includegraphics[height=0.7in]{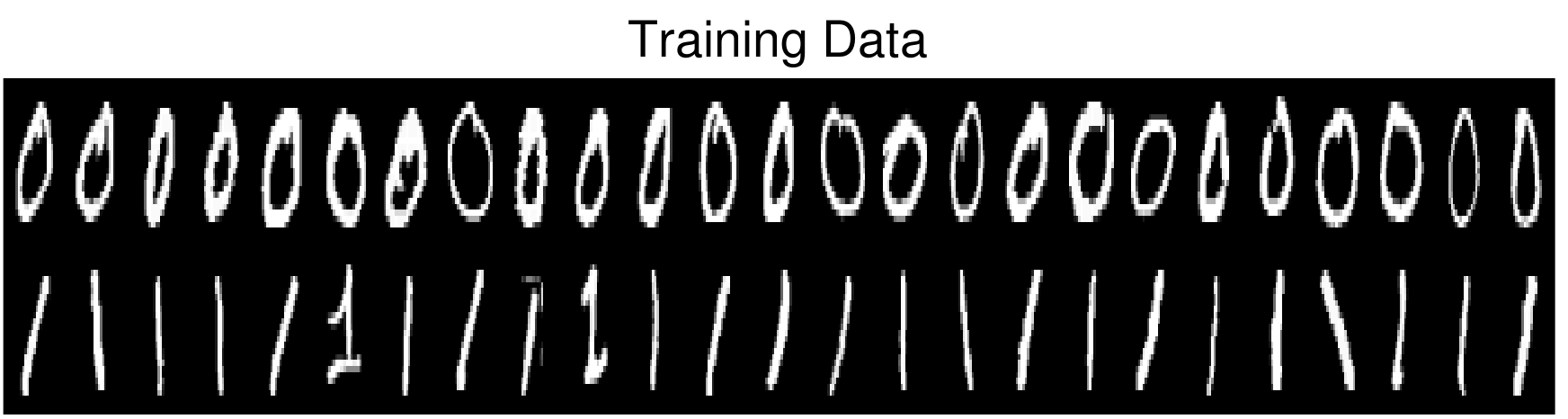}} &
\includegraphics[height=2in]{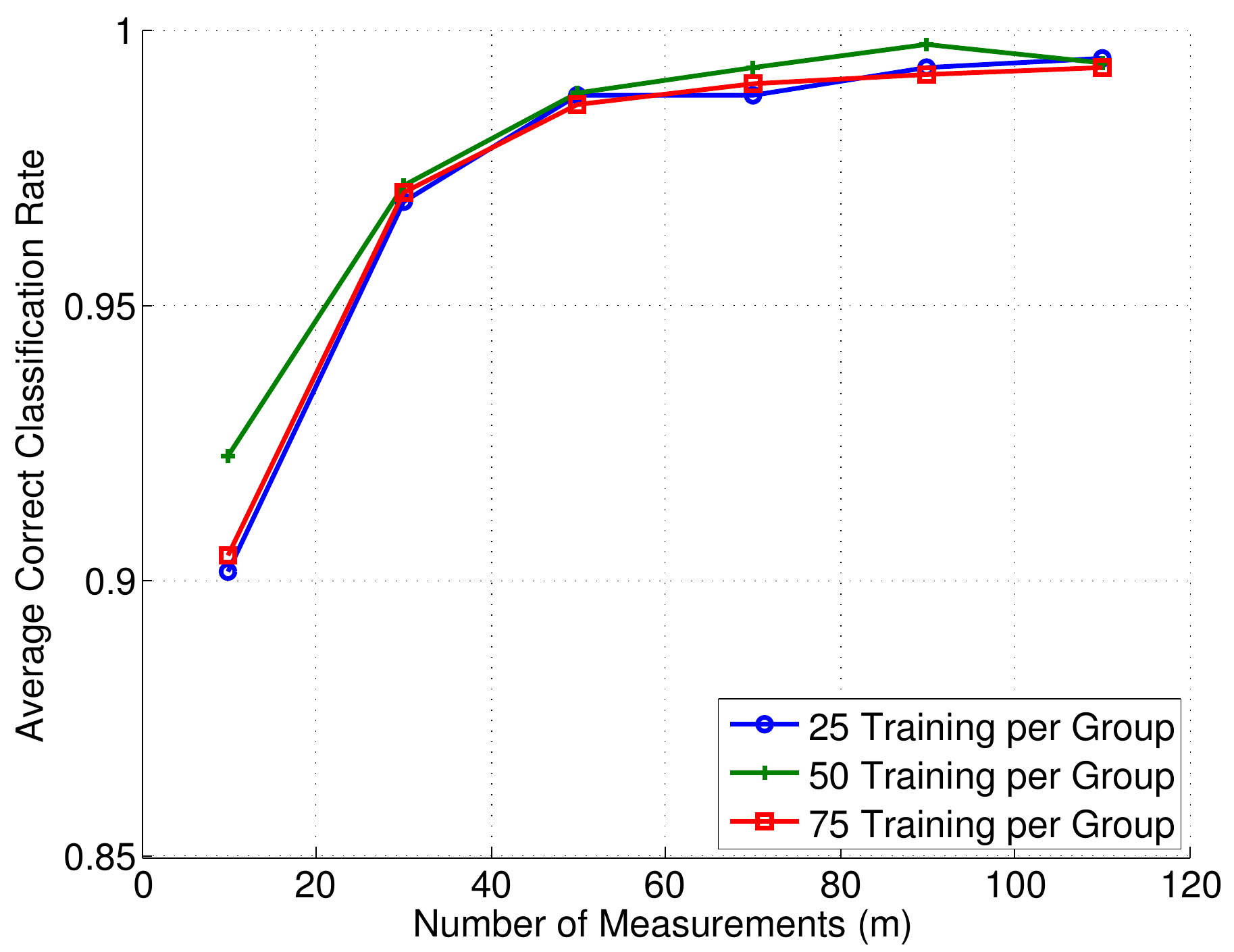} \\
\end{tabular}
\includegraphics[height=0.7in]{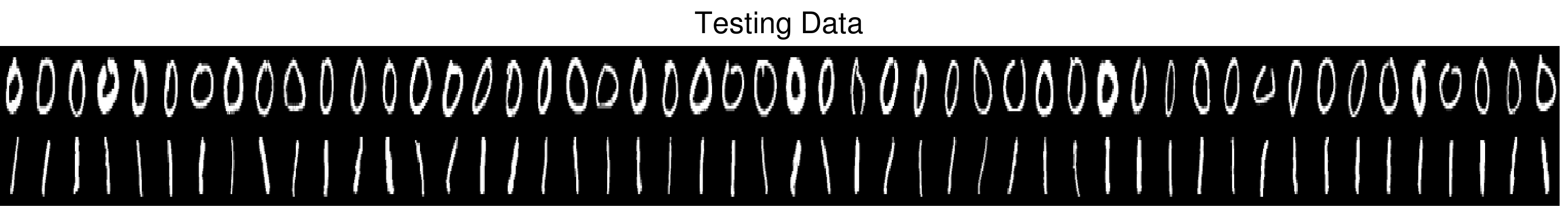} 
\caption{Classification experiment using the handwritten ``0" and ``1" digit images from the MNIST dataset, $L=1$, $n=28\times 28 =784$, 50 test points per group, and 30 trials of randomly generating $A$. (Top left) Training data images when $p = 50$. (Top right) Average correct classification rate versus $m$ and for the indicated number of training points per class. (Bottom) Testing data images.}
\label{mnist:01}
\end{figure}

First, we apply Algorithms \ref{proposed algorithm1} and \ref{proposed algorithm2} when considering only two digit classes. Figure \ref{mnist:01} shows the correct classification rate for the digits ``0" versus ``1". We set $m\in\{10,30,50,70,90,110\}$, $p\in\{50,100,150\}$ with equally sized training data sets for each class, and classify 50 images per digit class. Notice that the algorithm is performing very well for small $m$ in comparison to $n=28\times 28 =784$ and only a single layer. Figure \ref{mnist:01} shows the results of a similar setup for the digits ``0" and ``5". In this experiment, we increased  to four layers and achieve classification accuracy around $90\%$ at the high end of $m$ values tested. This indicates that the digits ``0" and ``5" are more likely to be mixed up than ``0" and ``1", which is understandable due to the more similar digit shape between ``0" and ``5". 

\begin{figure}[!htbp]
\centering
\includegraphics[height=2in]{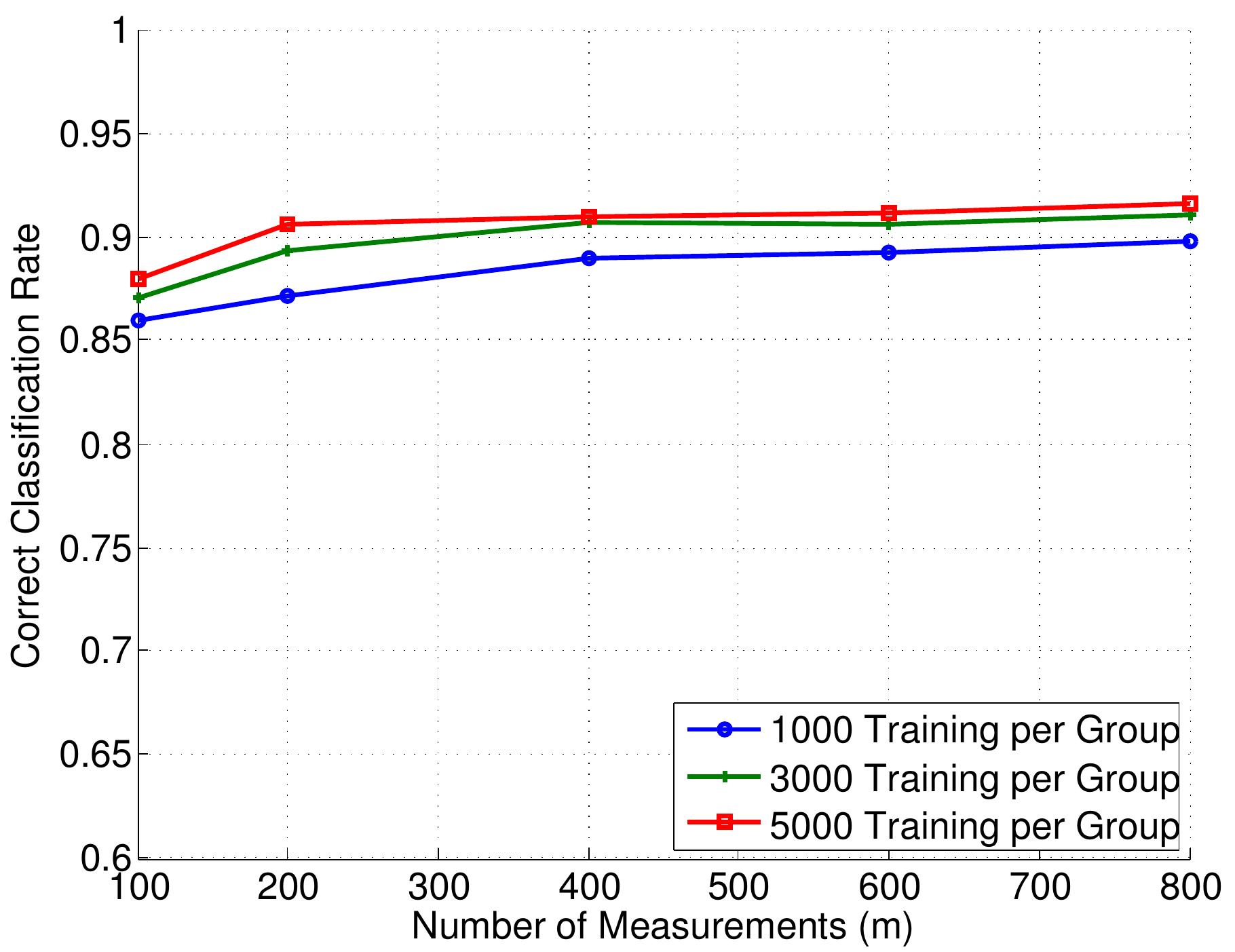}
\caption{Correct classification rate versus $m$ when using all ten (0-9) handwritten digits from the MNIST dataset, $L=18$, $n=28\times 28=784$, 1,000, 3,000, and 5,000 training points per group, 800 test points per group (8,000 total), and a single instance of randomly generating $A$.}
\label{mnist:all}
\end{figure}

Next, we apply Algorithms \ref{proposed algorithm1} and \ref{proposed algorithm2} to the MNIST dataset with all ten digits. We utilize $1,000$, $3,000$, and $5,000$ training points per digit class, and perform classification with $800$ test images per class. The classification results using 18 layers and $m\in\{100, 200, 400, 600, 800\}$ are shown in Figure \ref{mnist:all}, where it can be seen that with $5,000$ training points per class, above 90\% classification accuracy is achieved for $m\geq 200$. We also see that larger training sets result in slightly improved classification. 

\begin{figure}[!htbp]
\centering
\begin{tabular}{cc}
\raisebox{1.1\height}{\includegraphics[height=0.7in]{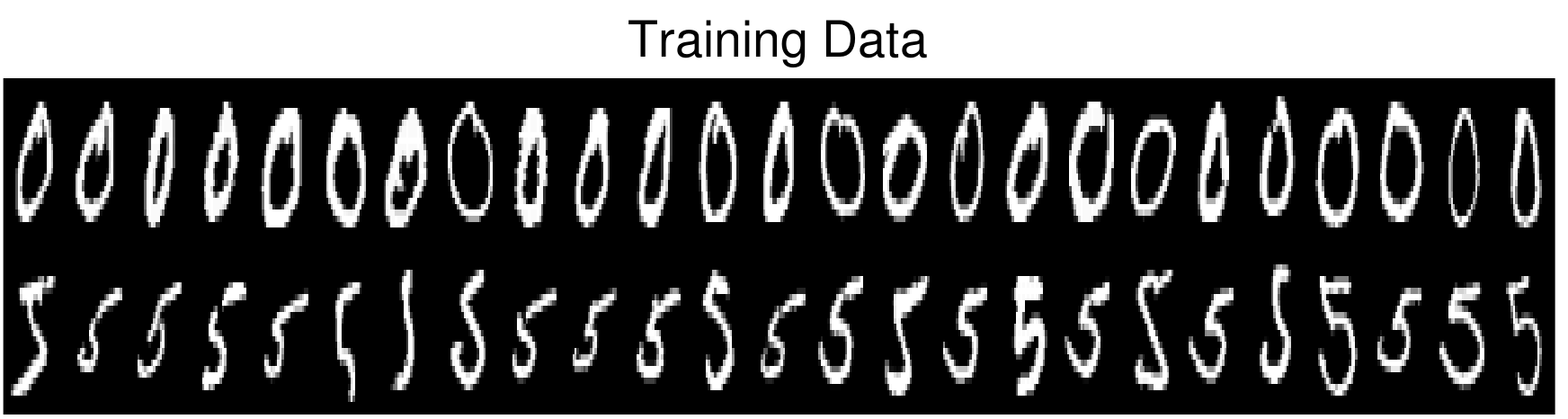}} &
\includegraphics[height=2in]{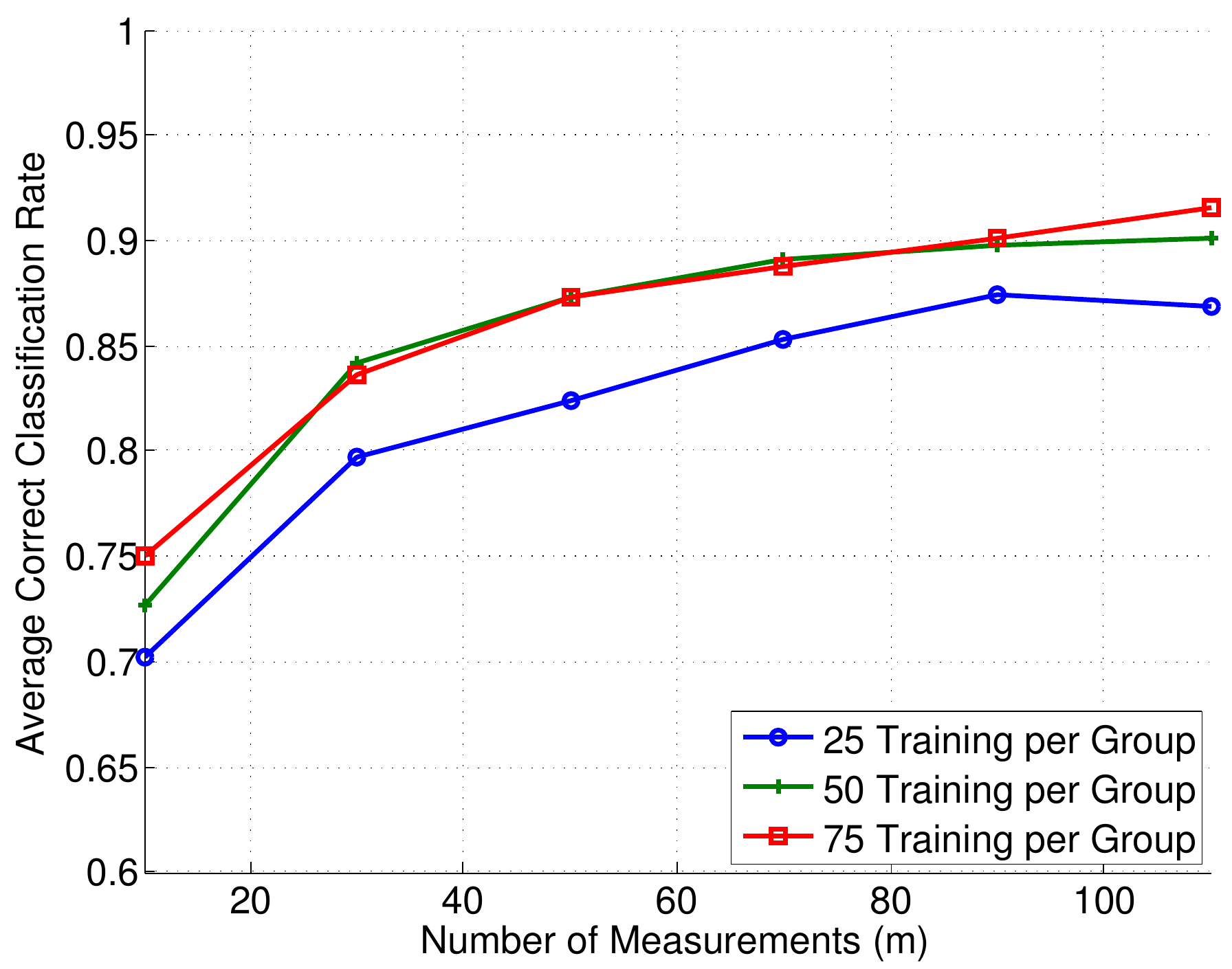}
\end{tabular}
\includegraphics[height=0.7in]{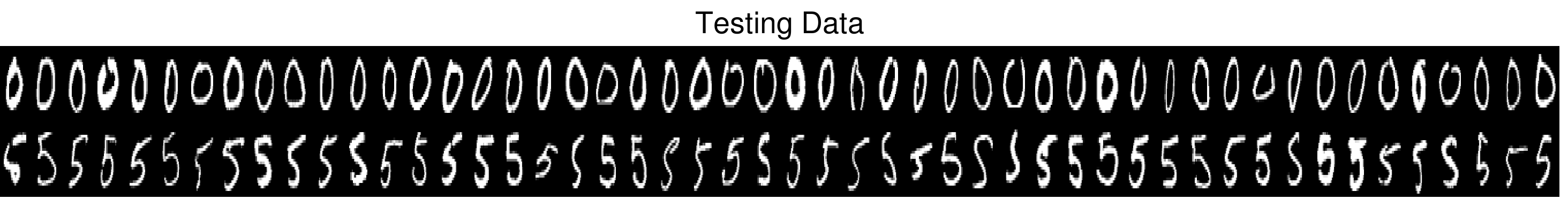} 
\caption{Classification experiment using the handwritten ``0" and ``5" digit images from the MNIST dataset, $L=4$, $n=28\times 28=784$, 50 test points per group, and 30 trials of randomly generating $A$. (Top left) Training data images when $p = 50$. (Top right) Average correct classification rate versus $m$ and for the indicated number of training points per class. (Bottom) Testing data images.}
\label{mnist:05}
\end{figure}

\clearpage
\subsection{Facial Recognition}

\begin{figure}[!htbp]
\centering
\begin{tabular}{cc}
\raisebox{1.1\height}{\includegraphics[height=0.7in,width=2.5in]{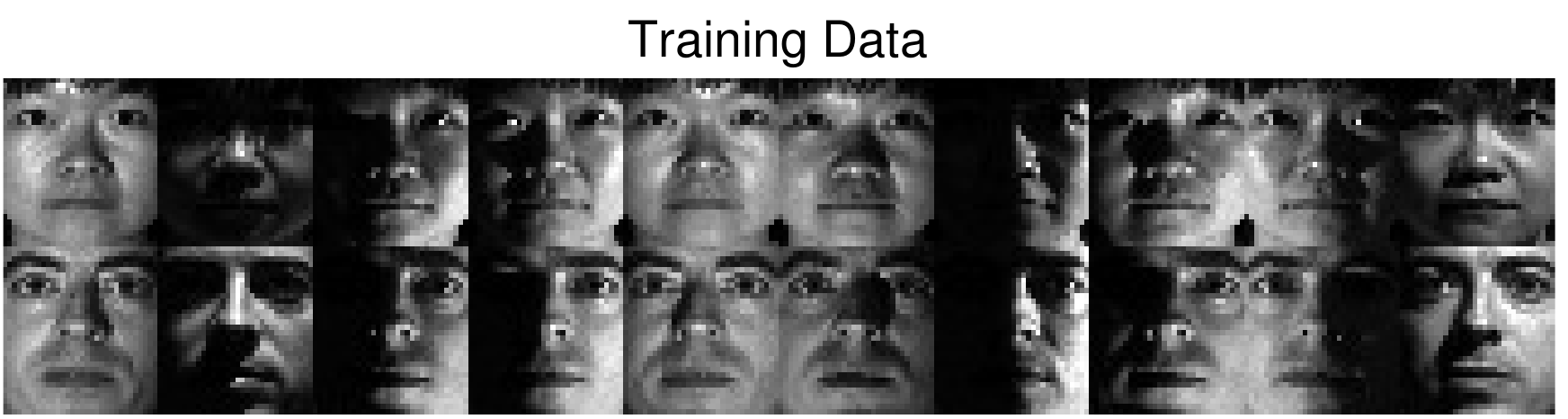}} &
\includegraphics[height=2in]{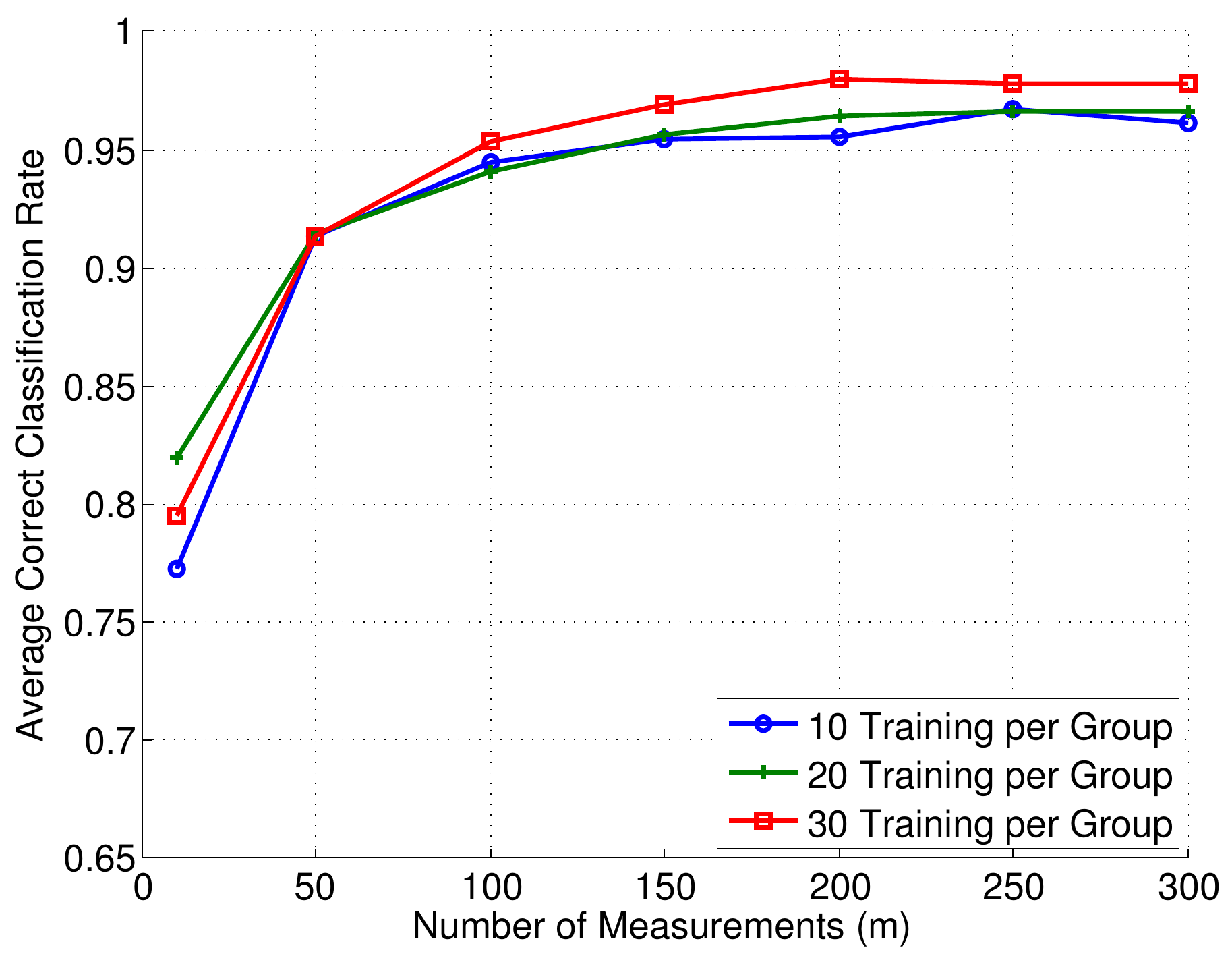} 
\end{tabular}
\includegraphics[height=0.7in]{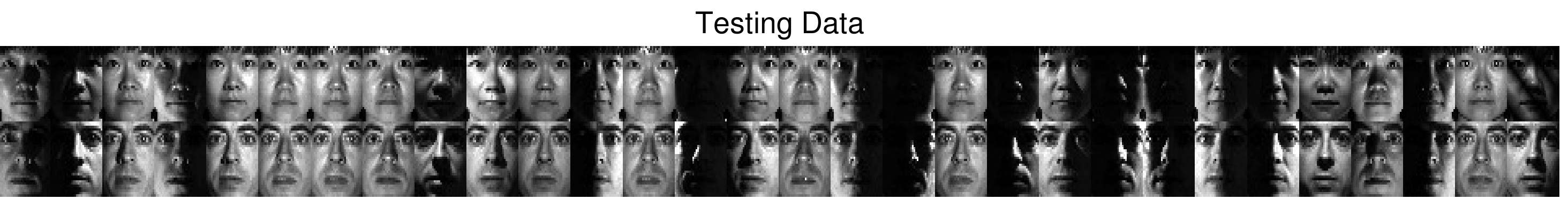} 
\caption{Classification experiment using two individuals from the extended YaleB dataset, $L=4$, $n=32\times 32 = 1024$, 30 test points per group, and 30 trials of randomly generating $A$. (Top left) Training data images when $p = 20$. (Top right) Average correct classification rate versus $m$ and for the indicated number of training points per class. (Bottom) Testing data images.}
\label{yaleB}
\end{figure}

Our last experiment considers facial recognition using the extended YaleB dataset \cite{CHHH07,CHH07b,CHHZ06,HYHNZ05}. This dataset  includes $32 \times 32$ images of 38 individuals with roughly 64 near-frontal images under different illuminations per individual. We select two individuals from the dataset, and randomly select images with different illuminations to be included in the training and testing sets (note that the same illumination was included for \textit{each} individual in the training and testing data). We execute Algorithms \ref{proposed algorithm1} and \ref{proposed algorithm2} using four layers with $m\in\{10,50,100,150,200,250,300 \}$, $p\in\{20,40,60\}$ with equally sized training data sets for each class, and classify 30 images per class. The results are displayed in Figure \ref{yaleB}. Above $95\%$ correct classification is achieved for $m\geq 150$ for each training set size included.

\section{Theoretical Analysis for a Simple Case}\label{section::theory}

\subsection{Main Results}
We now provide a theoretical analysis of Algorithms \ref{proposed algorithm1} and \ref{proposed algorithm2} in which we make a series of simplifying assumptions to make the development more tractable.
We focus on the setting where the signals are two-dimensional, belonging to one of two classes, and consider a single layer (i.e., $\ell=1$, $n=2$, and $G=2$). 
Moreover, we assume the true classes $G_1$ and $G_2$ to be two disjoint \textit{cones} in $\R^2$ and assume that regions of the same angular measure have the same number (or density) of training points.  We believe analyzing this setup will provide a foundation for a more generalized analysis in future work. 

\begin{figure}[!htbp]
\centering
\includegraphics[height=3in]{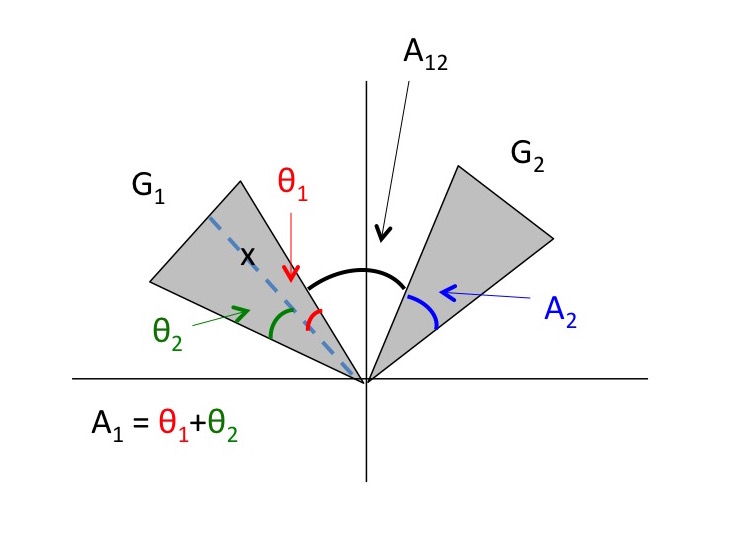}
\caption{Visualization of the analysis setup for two classes of two dimensions. If a hyperplane intersects the $\theta_1$ region of $G_1$, then $x$ is not on the same side of the hyperplane as $G_2$. If a hyperplane intersects the $\theta_2$ region of $G_1$, then $x$ is on the same side of the hyperplane as $G_2$. That is, $\theta_1$ and $\theta_2$ are determined by the position of $x$ within $G_1$, and $\theta_1+\theta_2 = A_1$.}
\label{2d cones}
\end{figure}

Let $A_1$ denote the angular measure of $G_1$, defined by \[A_1 = \max_{x_1, x_2\in G_1} \angle(x_1,x_2),\] where $\angle(x_1,x_2)$ denotes the angle between the vectors $x_1$ and $x_2$; define $A_2$ similarly for $G_2$.  Also, define \[A_{12} = \min_{x_1\in G_1, x_2\in G_2} \angle(x_1,x_2)\] as the angle between classes $G_1$ and $G_2$. Suppose that the test point $x\in G_1$, and  that we classify $x$ using $m$ random hyperplanes. For simplicity, we assume that the hyperplanes can intersect the cones, but only intersect \textit{one} cone at a time. This means we are imposing the condition $A_{12} + A_1 + A_2 \leq \pi$. See Figure \ref{2d cones} for a visualization of the setup for the analysis. Notice that $A_1$ is partitioned into two disjoint pieces, $\theta_1$ and $\theta_2$, where $A_1 = \theta_1+\theta_2$. The angles $\theta_1$ and $\theta_2$ are determined by the location of $x$ within $G_1$. 

The \RF parameter (\ref{RF3}) is still used; however, now we have angles instead of numbers of training points. That is,
\begin{align} \label{RF3 continuous}
r(\ell,i,t,g) &= \frac{A_{g|t}}{\sum_{j=1}^G A_{j|t}} \frac{\sum_{j=1}^G |A_{g|t} - A_{j|t}|}{\sum_{j=1}^G A_{j|t}},
\end{align}
where $A_{g|t}$ denotes the angle of class $g$ with the $t$-th sign pattern at the $i$-th set selection in the $\ell$-th layer.
Throughout, let $t_i^\star$ denote the sign pattern index of the test point $x$ with the $i$-th hyperplane at the first level (i.e., $\ell=1$). 
Letting $\widehat{b}_x$ denote the classification label for $x$ after running the proposed algorithm, Theorem \ref{main theorem} describes the probability that $x$ gets classified correctly with $\widehat{b}_x = 1$. Note that for simplicity, in Theorem \ref{main theorem} we assume the classes $G_1$ and $G_2$ are of the same size (i.e., $A_1=A_2$) and the test point $x$ lies in the middle of class $G_1$ (i.e., $\theta_1 = \theta_2$). These assumptions are for convenience and clarity of presentation only (note that \eqref{RF3 bound multinomial complement} is already quite cumbersome), but the proof follows analogously (albeit without easy simplifications) for the general case; for convenience we leave the computations in Table \ref{table::redness factors for 2d cones} in general form and do not utilize the assumption $\theta_1 = \theta_2$ until the end of the proof.

\begin{figure}[!htbp]
\centering
\includegraphics[height=3in]{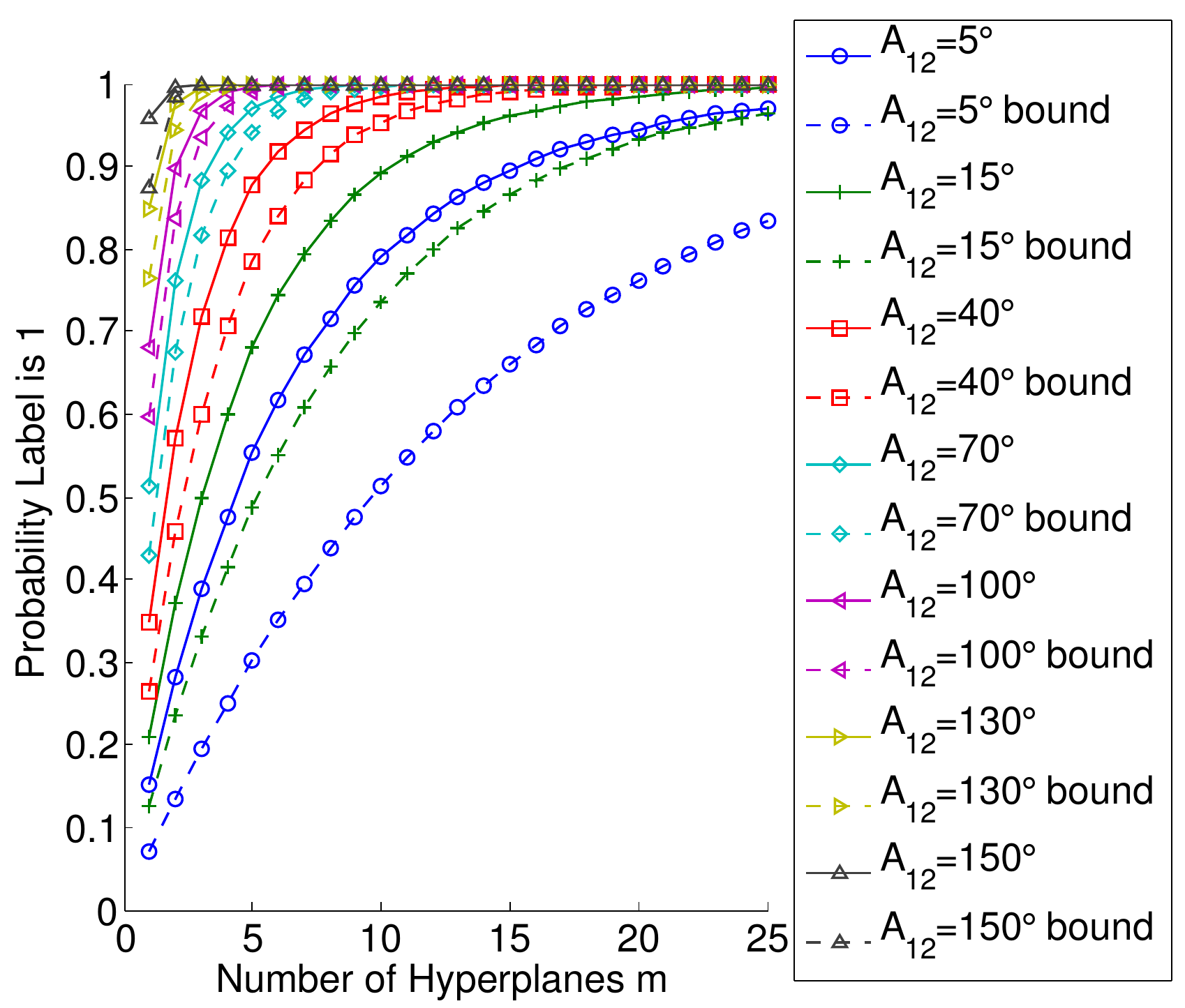}
\caption{$\mathbb{P}[\widehat{b}_x = 1]$ versus the number of hyperplanes $m$ when $A_{12}$ is varied (see legend), $A_1 = A_2 = 15^\circ$, and $\theta_1 = \theta_2 = 7.5^\circ$. The solid lines indicate the probability (\ref{classification with cutting continuous}) with the multinomial probability given by (\ref{multinomial probability}) and the conditional probability (\ref{conditional probability continuous RF3}) simulated over 1000 trials of the uniform random variables. The dashed lines indicate the result (\ref{RF3 bound multinomial complement}) provided in Theorem \ref{main theorem}.}
\label{theorem figure}
\end{figure}

\begin{theorem}\label{main theorem}
Let the classes $G_1$ and $G_2$ be two cones in $\R^2$ defined by angular measures $A_1$ and $A_2$, respectively, and suppose regions of the same angular measure have the same density of training points. Suppose $A_1 = A_2$, $\theta_1 = \theta_2$, and $A_{12} + A_1 + A_2 \leq \pi$. Then, the probability that a data point $x\in G_1$ gets classified in class $G_1$ by Algorithms \ref{proposed algorithm1} and \ref{proposed algorithm2} using a single layer and a measurement matrix $A\in\R^{m\times 2}$ with independent standard Gaussian entries is bounded as follows,
\begin{align}
\mathbb{P}[\widehat{b}_x = 1] &\geq 1 - \hspace{-8mm} \underset{j+k_{1,\theta_1}+k_{1,\theta_2}+k_2 + k = m, \,\, k_{1,\theta_2} \geq 9(j+k_{1,\theta_1})}{\sum_{j=0}^m \sum_{k_{1,\theta_1}=0}^m \sum_{k_{1,\theta_2}=0}^m \sum_{k_2=0}^m \sum_{k=0}^m} \binom{m}{j,k_{1,\theta_1},k_{1,\theta_2},k_2,k} \left(\frac{A_{12}}{\pi}\right)^j \left(\frac{A_1}{2\pi} \right)^{k_{1,\theta_1}+k_{1,\theta_2}}   \notag \\
&\quad\quad\times \left(\frac{A_1}{\pi} \right)^{k_2} \left(\frac{\pi-2A_1-A_{12}}{\pi}\right)^k. \label{RF3 bound multinomial complement}
\end{align}
\end{theorem}
Figure \ref{theorem figure} displays the classification probability bound of Theorem \ref{main theorem} compared to the (simulated) true value of $\mathbb{P}[\widehat{b}_x = 1]$. 
Here, $A_1 = A_2 = 15^\circ$, $\theta_1 = \theta_2 = 7.5^\circ$, and $A_{12}$ and $m$ are varied. Most importantly, notice that in all cases, the classification probability is approaching 1 with increasing $m$. Also, the result from Theorem \ref{main theorem} behaves similarly as the simulated true probability, especially as $m$ and $A_{12}$ increase.

The following two corollaries provide asymptotic results for situations where $\mathbb{P}[\widehat{b}_x = 1]$ tends to 1 when $m\rightarrow\infty$. Corollary \ref{Corollary 1} provides this result whenever $A_{12}$ is at least as large as both $A_1$ and $\pi-2A_1-A_{12}$, and Corollary \ref{Corollary 2} provides this result for certain combinations of $A_1$ and $A_{12}$.

\begin{cor}\label{Corollary 1}
Consider the setup of Theorem \ref{main theorem}. Suppose $A_{12} \geq A_1$ and $A_{12} \geq \pi - 2A_1 - A_{12}$. Then $\mathbb{P}[\widehat{b}_x = 1] \rightarrow 1$ as $m\rightarrow \infty$.
\end{cor}

\begin{cor}\label{Corollary 2}
Consider the setup of Theorem \ref{main theorem}. Suppose $A_1 + A_{12} > 0.58 \pi$ and $A_{12} + \frac{3}{4}A_1 \leq \frac{\pi}{2}$. Then $\mathbb{P}[\widehat{b}_x = 1] \rightarrow 1$ as $m\rightarrow \infty$.
\end{cor}
\begin{remark}
Note that the two conditions in Corollary \ref{Corollary 2} imply the assumption that $A_1 \geq 0.32\pi$ and $A_{12} \leq 0.26 \pi$.
\end{remark}

\subsection{Proof of Main Results}

\subsubsection{Proof of Theorem \ref{main theorem}}

\begin{proof}
Using our setup, we have five possibilities for any given hyperplane: (i) the hyperplane completely separates the two classes, i.e., the cones associated with the two classes fall on either side of the hyperplane, (ii) the hyperplane completely does not separate the two classes, i.e.,  the cones fall on the same side of the hyperplane, (iii) the hyperplane cuts through $G_2$, (iv) the hyperplane cuts through $G_1$ via $\theta_1$, or (v) the hyperplane cuts through $G_1$ via $\theta_2$. Using this observation, we can now define the event 
\begin{equation}\label{theevent}
E( j,k_{1,\theta_1},k_{1,\theta_2},k_2 )
\end{equation} 
whereby from among the $m$ total hyperplanes, $j$ 
hyperplanes  separate the cones, $k_{1,\theta_1}$ hyperplanes  cut  $G_1$ in  $\theta_1$, $k_{1,\theta_2}$ hyperplanes cut  $G_1$  in  $\theta_2$, and  $k_2$ hyperplanes cut  $G_2$. 
See Table \ref{table::redness factors for 2d cones} for an easy reference of these quantities. Note that we must distinguish between hyperplanes that cut through $\theta_1$ and those that cut through $\theta_2$; $k_{1,\theta_1}$ hyperplanes cut $G_1$ and land within $\theta_1$ so that $x$ is \textit{not} on the same side of the hyperplane as $G_2$ whereas $k_{1,\theta_2}$ hyperplanes cut $G_1$ and land within $\theta_2$ so that $x$ \textit{is} on the same side of the hyperplane as $G_2$. These orientations will affect the computation of the membership index. Using the above definition of (\ref{theevent}), we use the law of total probability to get a handle on $\mathbb{P}[\widehat{b}_x = 1]$, the probability that the test point $x$ gets classified correctly, as follows,

{\small
\begin{align}
\mathbb{P}[\widehat{b}_x = 1] &= \mathbb{P}\left[\sum_{i=1}^m r(\ell,i,t_i^\star,1) > \sum_{i=1}^m r(\ell,i,t_i^\star,2)\right] \notag\\
&= \underset{j+k_{1,\theta_1}+k_{1,\theta_2}+k_2\leq m}{\sum_{j,k_{1,\theta_1}, k_{1,\theta_2},k_2 } }
\mathbb{P}\left[\sum_{i=1}^m r(\ell,i,t^\star_i,1) > \sum_{i=1}^m r(\ell,i,t^\star_i,2) \;| E( j,k_{1,\theta_1},k_{1,\theta_2},k_2 )
\right]\notag\\
&\qquad\qquad\qquad\qquad\times \mathbb{P}\left[
E( j,k_{1,\theta_1},k_{1,\theta_2},k_2 )
\right]. \label{classification with cutting continuous}
\end{align}}
 The latter probability in (\ref{classification with cutting continuous}) is similar to the probability density of a multinomial random variable:
\begin{align}
&\mathbb{P}\left[E( j,k_{1,\theta_1},k_{1,\theta_2},k_2 )
\right] \notag \\
&= \binom{m}{j,k_{1,\theta_1},k_{1,\theta_2},k_2,m-j-k_{1,\theta_1}-k_{1,\theta_2}-k_2} \left(\frac{A_{12}}{\pi}\right)^j \left(\frac{\theta_1}{\pi} \right)^{k_{1,\theta_1}} \left(\frac{\theta_2}{\pi} \right)^{k_{1,\theta_2}} \notag \\
&\quad\times \left(\frac{A_2}{\pi} \right)^{k_2} \left(\frac{\pi-A_1-A_2-A_{12}}{\pi}\right)^{m-j-k_{1,\theta_1}-k_{1,\theta_2}-k_2}, \label{multinomial probability}
\end{align}
where $\binom{n}{k_1,k_2,\dots,k_m} = \frac{n!}{k_1!k_2!\cdots k_m!}$.

To evaluate the conditional probability in (\ref{classification with cutting continuous}), we must determine the value of $r(\ell,i,t_i^\star,g)$, for $g=1,2$, given the hyperplane cutting pattern event. Table \ref{table::redness factors for 2d cones} summarizes the possible cases. In the cases where the hyperplane cuts through either $G_1$ or $G_2$, we model the location of the hyperplane within the class by a uniform random variable. We will use the notation $z\sim U(a,b)$ to indicate that $z$ is a uniform random variable over the interval $[a,b]$, and let $u$, $u'$, $u_h$, $u_h'$ (for an index $h$) denote independent copies of a $U(0,1)$ uniform random variable; therefore $\theta u \sim U(0, \theta)$. 

\begin{table}[!htbp]
\centering
\begin{tabular}{| c | c |c | c |}
\hline
Hyperplane Case & Number in event \eqref{theevent} & Class $g$ & Value of $r(\ell,i,t_i^\star,g)$ (see \eqref{RF3 continuous})\\ 
\hline
\hline
\multirow{2}{*}{(i) separates} & \multirow{2}{*}{$j$} & 1 & $1$\\ 
 & & 2 & $0$\\ 
\hline
\hline
\multirow{2}{*}{(ii) does not separate} & \multirow{2}{*}{$m - j - k_2 - k_{1,\theta_1} - k_{1,\theta_2}$}  &  1 & $\frac{A_1|A_1-A_2|}{(A_1+A_2)^2}$ \\ 
&& 2 & $\frac{A_2|A_1-A_2|}{(A_1+A_2)^2}$\\ 
\hline
\hline
\multirow{2}{*}{(iii) cuts $G_2$} & \multirow{2}{*}{$k_2$} &  1 & $\frac{A_1|A_1-A_2u'|}{(A_1+A_2u')^2}$\\ 
&& 2 & $\frac{A_2 u'|A_1-A_2u'|}{(A_1+A_2u')^2}$\\ 
\hline
\hline
\multirow{2}{*}{(iv) cuts $G_1$, $\theta_1$}& \multirow{2}{*}{$k_{1,\theta_1}$} & 1 & $1$\\ 
&& 2 & $0$\\ 
\hline
\hline
\multirow{2}{*}{(v) cuts $G_1$, $\theta_2$} & \multirow{2}{*}{$k_{1,\theta_2}$} & 1 & $\frac{(\theta_1+\theta_2u)|\theta_1+\theta_2u-A_2|}{(\theta_1+\theta_2u+A_2)^2}$ \\ 
&& 2 & $\frac{A_2|\theta_1+\theta_2u-A_2|}{(\theta_1+\theta_2u+A_2)^2}$ \\ 
\hline
\end{tabular}
\caption{Summary of (\ref{RF3 continuous}) when up to one cone can be cut per hyperplane, where $ u, u'$ are indepdendent $U(0,1)$ random variables.} 
\label{table::redness factors for 2d cones}
\end{table}

Using the computations given in Table \ref{table::redness factors for 2d cones} and assuming $j$ hyperplanes separate (i.e. condition (i) described above), $k_{1,\theta_1}$ hyperplanes cut $G_1$ in $\theta_1$ (condition (iv) above), $k_{1,\theta_2}$ hyperplanes cut $G_1$ in $\theta_2$ (condition (v) above),  $k_2$ hyperplanes cut $G_2$ (condition (iii) above), and $m-j-k_{1,\theta_1}-k_{1,\theta_2}-k_2$ hyperplanes do not separate (condition (ii) above), we compute the membership index parameters defined in \eqref{RF3 continuous} as: 
\begin{align}\label{compRF1}
\sum_{i=1}^m r(\ell,i,t_i^\star,1) &= j + (m-j-k_{1,\theta_1}-k_{1,\theta_2}-k_2)\frac{A_1|A_1-A_2|}{(A_1+A_2)^2} + k_{1,\theta_1} \notag \\
&\quad + \sum_{h=1}^{k_{1,\theta_2}}\frac{(\theta_1+\theta_2 u_h)|\theta_1+\theta_2 u_h-A_2|}{(\theta_1+\theta_2 u_h+A_2)^2} + \sum_{h=1}^{k_2}\frac{A_1|A_1-A_2 u_h'|}{(A_1+A_2 u_h')^2} \notag\\
&=  j + k_{1,\theta_1} 
 + \sum_{h=1}^{k_{1,\theta_2}}\frac{(\theta_1+\theta_2 u_h)|\theta_1+\theta_2 u_h-A_1|}{(\theta_1+\theta_2 u_h+A_1)^2} + \sum_{h=1}^{k_2}\frac{A_1|A_1-A_1 u_h'|}{(A_1+A_1 u_h')^2}
\end{align}
and
\begin{align}\label{compRF2}
\sum_{i=1}^m r(\ell,i,t_i^\star,2) &=  (m-j-k_{1,\theta_1}-k_{1,\theta_2}-k_2)\frac{A_2|A_1-A_2|}{(A_1+A_2)^2}  \notag \\
&\quad + \sum_{h=1}^{k_{1,\theta_2}}\frac{A_2|\theta_1+\theta_2 u_h-A_2|}{(\theta_1+\theta_2 u_h+A_2)^2} + \sum_{h=1}^{k_2}\frac{A_2 u_h'|A_1-A_2 u_h'|}{(A_1+A_2 u_h')^2}\notag\\
&=  \sum_{h=1}^{k_{1,\theta_2}}\frac{A_1|\theta_1+\theta_2 u_h-A_1|}{(\theta_1+\theta_2 u_h+A_1)^2} + \sum_{h=1}^{k_2}\frac{A_1 u_h'|A_1-A_1 u_h'|}{(A_1+A_1 u_h')^2},
\end{align}
where in both cases we have simplified using the assumption $A_1 = A_2$.
Thus, the conditional probability in (\ref{classification with cutting continuous}), can be expressed as:
\begin{align}
&\mathbb{P}\left[j + k_{1,\theta_1} + \sum_{h=1}^{k_{1,\theta_2}}\frac{|\theta_1+\theta_2 u_h-A_1|(\theta_1+\theta_2 u_h-A_1)}{(\theta_1+\theta_2 u_h+A_1)^2} + \sum_{h=1}^{k_2}\frac{|A_1-A_1 u_h'|(A_1-A_1 u_h')}{(A_1+A_1 u_h')^2} > 0\right] \label{conditional probability continuous RF3},
\end{align}
where it is implied that this probably is conditioned on the hyperplane configuration as in \eqref{classification with cutting continuous}.
Once the probability (\ref{conditional probability continuous RF3}) is known, we can calculate the full classification probability (\ref{classification with cutting continuous}).

Since by assumption, $\theta_1 + \theta_2 = A_1$, we have $\theta_1+\theta_2 u-A_1 \leq  0 $ and $A_1 -A_1 u' \geq  0$. Thus,  (\ref{conditional probability continuous RF3}) simplifies to
\begin{align}
&\mathbb{P}\left[j + k_{1,\theta_1} - \sum_{h=1}^{k_{1,\theta_2}}\frac{(\theta_1+\theta_2 u_h-A_1)^2}{(\theta_1+\theta_2 u_h+A_1)^2} + \sum_{h=1}^{k_2}\frac{(A_1-A_1 u_h')^2}{(A_1+A_1 u_h')^2} > 0\right]\notag\\
&= \mathbb{P}\left[j + k_{1,\theta_1} - \sum_{h=1}^{k_{1,\theta_2}}\frac{(\theta_2 u_h)^2}{(\theta_1+\theta_2 u_h+A_1)^2} + \sum_{h=1}^{k_2}\frac{(A_1 u_h')^2}{(A_1+A_1 u_h')^2} > 0\right]. \label{conditional probability continuous RF3 A1=A2}
\end{align}
Next, using that $\theta_2 u \geq 0$ and $A_1 u' \leq A_1$ for the random variables in the denominators, we can bound (\ref{conditional probability continuous RF3 A1=A2}) from below by 
\begin{align}
(\ref{conditional probability continuous RF3 A1=A2}) \geq \mathbb{P}\left[j + k_{1,\theta_1} - \sum_{h=1}^{k_{1,\theta_2}}\frac{(\theta_2 u_h)^2}{(\theta_1+A_1)^2} + \sum_{h=1}^{k_2}\frac{(A_1 u_h')^2}{(2A_1)^2} > 0\right]. \label{RF3 first bound}
\end{align}
Letting $\theta>0$ to be chosen later (and abusing notation slightly to allow $u$, $u'$ to be new independent uniform random variables), we can rewrite the above probability (\ref{RF3 first bound}) as
\begin{align}
P:= &\mathbb{P}\left[j + k_{1,\theta_1} - \frac{1}{(\theta_1+A_2)^2}\sum_{h=1}^{k_{1,\theta_2}}(\theta_2u_h)^2 + \frac{1}{(2A_1)^2} \sum_{h=1}^{k_2} (A_1u_h')^2> 0\right] \notag \\
&= \mathbb{P}\left[ \frac{1}{(\theta_1+A_1)^2}\sum_{h=1}^{k_{1,\theta_2}}(\theta_2 u_h)^2 - \frac{1}{(2A_1)^2} \sum_{h=1}^{k_2} (A_1u_h')^2<  j + k_{1,\theta_1}\right] \notag \\
&= 1- \mathbb{P}\left[ \frac{1}{(\theta_1+A_1)^2}\sum_{h=1}^{k_{1,\theta_2}}(\theta_2 u_h)^2 - \frac{1}{(2A_1)^2} \sum_{h=1}^{k_2} (A_1 u_h')^2 \geq  j + k_{1,\theta_1}\right] \notag \\
&= 1- \mathbb{P}\left[ e^{\theta \left(\frac{1}{(\theta_1+A_1)^2}\sum_{h=1}^{k_{1,\theta_2}}(\theta_2 u_h)^2 - \frac{1}{(2A_1)^2} \sum_{h=1}^{k_2} (A_1 u_h')^2\right)} \geq  e^{\theta(j + k_{1,\theta_1})}\right] \notag \\
&\geq 1-e^{-\theta(j + k_{1,\theta_1})} \mathbb{E}\left[ e^{\theta \left(\frac{1}{(\theta_1+A_1)^2}\sum_{h=1}^{k_{1,\theta_2}}(\theta_2 u_h)^2 - \frac{1}{(2A_1)^2} \sum_{h=1}^{k_2} (A_1 u_h')^2\right)} \right] \label{Markov step}\end{align}
where in the last equality, $\theta$ is a non-negative parameter to be chosen later, and (\ref{Markov step}) follows from Markov's inequality. Continuing and using the independence of each hyperplane, we now have 

\begin{align}
P &\geq 1-e^{-\theta(j + k_{1,\theta_1})} \mathbb{E}\left[ \prod_{h=1}^{k_{1,\theta_2}} e^{ \frac{\theta}{(\theta_1+A_1)^2}(\theta_2 u_h)^2}\right] \mathbb{E}\left[ \prod_{h=1}^{k_2} e^{- \frac{\theta}{(2A_1)^2} (A_1 u_h')^2)} \right] \notag \\
&= 1-e^{-\theta(j + k_{1,\theta_1})}\prod_{h=1}^{k_{1,\theta_2}} \mathbb{E}\left[ e^{ \frac{\theta}{(\theta_1+A_1)^2}(\theta_2 u_h)^2}\right] \prod_{h=1}^{k_2} \mathbb{E}\left[ e^{- \frac{\theta}{(2A_1)^2} (A_1 u_h')^2)} \right] \notag \\
&= 1-e^{-\theta(j + k_{1,\theta_1})}\left(\mathbb{E}\left[ e^{ \frac{\theta}{(\theta_1+A_1)^2}(\theta_2 u)^2}\right] \right)^{k_{1,\theta_2}} \left( \mathbb{E}\left[ e^{- \frac{\theta}{(2A_1)^2} (A_1 u')^2)} \right]\right)^{k_2}. \label{last equality}
\end{align}
 Next, one readily computes the following (we include the computation in the appendix, Section \ref{app:erf}, for completeness):
\begin{align}\label{helloerf}
\mathbb{E}\left[ e^{ \frac{\theta}{(\theta_1+A_1)^2}(\theta_2 u)^2}\right] = \frac{\sqrt{\pi} \mbox{erfi}(\frac{\theta_2}{A_1+\theta_1}\sqrt{\theta})}{\frac{\theta_2}{A_1+\theta_1}\sqrt{\theta}}
\end{align}
and
\begin{align}\label{helloerf2}
\mathbb{E}\left[ e^{- \frac{\theta}{(2A_1)^2} (A_1 u')^2)} \right] = \frac{\sqrt{\pi}\mbox{erf}(\frac{1}{2}\sqrt{\theta})}{\frac{1}{2}\sqrt{\theta}},
\end{align}
where $\mbox{erf}(x) = \frac{2}{\sqrt{\pi}} \int_0^x e^{-t^2} \,\, dt$ is the error function and $\mbox{erfi}(x) = -i\mbox{erf}(ix)= \frac{2}{\sqrt{\pi}} \int_0^x e^{t^2} \,\, dt$ is the imaginary error function. 
Therefore,
\begin{align}
&P \geq 1-e^{-\theta(j + k_{1,\theta_1})}\left(\frac{\sqrt{\pi} \mbox{erfi}(\frac{\theta_2}{A_1+\theta_1}\sqrt{\theta})}{\frac{\theta_2}{A_1+\theta_1}\sqrt{\theta}} \right)^{k_{1,\theta_2}} \left( \frac{\sqrt{\pi}\mbox{erf}(\frac{1}{2}\sqrt{\theta})}{\frac{1}{2}\sqrt{\theta}}\right)^{k_2}. \label{RF3 bound erf}
\end{align}
Now, we note the following trivial upper bounds on the $\mbox{erf}$ and $\mbox{erfi}$ functions.
\begin{align}
\mbox{erf}(x) = \frac{2}{\sqrt{\pi}} \int_0^x e^{-t^2}\,\,dt \leq \frac{2}{\sqrt{\pi}} \int_0^x 1 \,\,dt = \frac{2x}{\sqrt{\pi}} \\
\mbox{erfi}(x) = \frac{2}{\sqrt{\pi}} \int_0^x e^{t^2} \,\,dt \leq \frac{2}{\sqrt{\pi}} e^{x^2} \int_0^x 1 \,\, dt = \frac{2x}{\sqrt{\pi}} e^{x^2}.
\end{align}
Applying the above bounds to \eqref{RF3 bound erf} 
gives 

\begin{align}
\mathbb{P}[\widehat{b}_x = 1] &\geq 
1-e^{-\theta(j + k_{1,\theta_1})}\left(\frac{\sqrt{\pi} \mbox{erfi}(\frac{\theta_2}{A_1+\theta_1}\sqrt{\theta})}{\frac{\theta_2}{A_1+\theta_1}\sqrt{\theta}} \right)^{k_{1,\theta_2}} \left( \frac{\sqrt{\pi}\mbox{erf}(\frac{\sqrt{\theta}}{2})}{\frac{\sqrt{\theta}}{2}}\right)^{k_2} \notag\\
&\geq 1-2^{k_{1,\theta_2}+k_2} e^{-\theta(j+k_{1,\theta_1})} e^{k_{1,\theta_2}\theta(\frac{\theta_2}{A_1+\theta_1})^2}\notag\\
&= 1-2^{k_{1,\theta_2}+k_2} e^{\theta\left(k_{1,\theta_2}(\frac{\theta_2}{A_1+\theta_1})^2 -(j+k_{1,\theta_1})\right)}\notag\\
&= 1-\alpha e^{\theta(\beta-\gamma)} =: f(\theta) \label{RF3 bound erf trivial}
\end{align}
where $$\alpha = 2^{k_{1,\theta_2}+k_2},\quad\beta = k_{1,\theta_2}\left(\frac{\theta_2}{A_1+\theta_1}\right)^2\quad \text{and} \quad\gamma = j+k_{1,\theta_1}.$$ Recall that we wish to choose $\theta>0$ such that $f(\theta)$ in \eqref{RF3 bound erf trivial} is maximized. If $\beta-\gamma <0$, then taking $\theta\rightarrow\infty$ maximizes $f(\theta)$ as $f(\theta)\rightarrow 1$. If $\beta-\gamma\geq 0$, then taking $\theta\rightarrow 0$ maximizes $f(\theta)$ as $f(\theta)\rightarrow 1-\alpha < 0$. But, since we are bounding a probability, we always have the trivial lower bound of zero. So, when $\beta-\gamma \geq 0$ we can use the simple bound $ \mathbb{P}[\widehat{b}_x = 1] \geq 0$.

Therefore, the probability of interest \eqref{classification with cutting continuous} reduces to (note the bounds on the summation indices):
\begin{align}
\mathbb{P}[\widehat{b}_x = 1] 
&= \underset{j+k_{1,\theta_1}+k_{1,\theta_2}+k_2\leq m}{\sum_{j,k_{1,\theta_1}, k_{1,\theta_2},k_2 } }
\mathbb{P}\left[\sum_{i=1}^m r(\ell,i,t^\star_i,1) > \sum_{i=1}^m r(\ell,i,t^\star_i,2) \;| E( j,k_{1,\theta_1},k_{1,\theta_2},k_2 )
\right]\notag\\
&\qquad\qquad\qquad\qquad\times \mathbb{P}\left[
E( j,k_{1,\theta_1},k_{1,\theta_2},k_2 )
\right]\\
&\geq  
{\sum_{\substack{j,k_{1,\theta_1}, k_{1,\theta_2},k_2 \\ j+k_{1,\theta_1}+k_{1,\theta_2}+k_2\leq m, \\ \beta-\gamma<0} }}
 \binom{m}{j,k_{1,\theta_1},k_{1,\theta_2},k_2,m-j-k_{1,\theta_1}-k_{1,\theta_2}-k_2} \left(\frac{A_{12}}{\pi}\right)^j \left(\frac{\theta_1}{\pi} \right)^{k_{1,\theta_1}} \notag\\
&\quad\quad \times  \left(\frac{\theta_2}{\pi} \right)^{k_{1,\theta_2}} \left(\frac{A_2}{\pi} \right)^{k_2} \left(\frac{\pi-A_1-A_2-A_{12}}{\pi}\right)^{m-j-k_{1,\theta_1}-k_{1,\theta_2}-k_2}.\label{choose theta bound}
\end{align}
The condition $\beta-\gamma < 0$ is equivalent to $k_{1,\theta_2}(\frac{\theta_2}{A_1+\theta_1})^2 - ( j+k_{1,\theta_1}) < 0$, which implies $ k_{1,\theta_2}(\frac{\theta_2}{A_1+\theta_1})^2 < j+k_{1,\theta_1}$. Assuming $\theta_1=\theta_2$ simplifies this condition to depend \textit{only} on the hyperplane configuration (and not $A_1$, $\theta_1$, and $\theta_2$) since $\frac{\theta_2}{A_1+\theta_1} = \frac{\theta_2}{3\theta_2} = \frac{1}{3}$. Thus, the condition $\beta-\gamma <0$ reduces to the condition $k_{1,\theta_2} < 9(j+k_{1,\theta_1})$ 
and (\ref{choose theta bound}) then simplifies to
\begin{align}
&
{\sum_{\substack{j+k_{1,\theta_1}+k_{1,\theta_2}+k_2\leq m, \\ k_{1,\theta_2} < 9(j+k_{1,\theta_1})}} 
}
 \binom{m}{j,k_{1,\theta_1},k_{1,\theta_2},k_2,m-j-k_{1,\theta_1}-k_{1,\theta_2}-k_2} \left(\frac{A_{12}}{\pi}\right)^j \left(\frac{\theta_1}{\pi} \right)^{k_{1,\theta_1}+k_{1,\theta_2}} \notag\\
&\quad\quad \times  \left(\frac{A_2}{\pi} \right)^{k_2} \left(\frac{\pi-2A_1-A_{12}}{\pi}\right)^{m-j-k_{1,\theta_1}-k_{1,\theta_2}-k_2} \\
&= \sum_{\substack{j+k_{1,\theta_1}+k_{1,\theta_2}+k_2 + k = m, \\ k_{1,\theta_2} < 9(j+k_{1,\theta_1})}}
\binom{m}{j,k_{1,\theta_1},k_{1,\theta_2},k_2,k} \left(\frac{A_{12}}{\pi}\right)^j \left(\frac{\theta_1}{\pi} \right)^{k_{1,\theta_1}+k_{1,\theta_2}}   \left(\frac{A_2}{\pi} \right)^{k_2} \left(\frac{\pi-2A_1-A_{12}}{\pi}\right)^k, \\
&= 
\sum_{\substack{j+k_{1,\theta_1}+k_{1,\theta_2}+k_2 + k = m, \\ k_{1,\theta_2} < 9(j+k_{1,\theta_1})}}\binom{m}{j,k_{1,\theta_1},k_{1,\theta_2},k_2,k} \left(\frac{A_{12}}{\pi}\right)^j \left(\frac{A_1}{2\pi} \right)^{k_{1,\theta_1}+k_{1,\theta_2}}   \left(\frac{A_1}{\pi} \right)^{k_2} \left(\frac{\pi-2A_1-A_{12}}{\pi}\right)^k, \label{RF3 bound multinomial}
\end{align}
where we have introduced $k$ to denote the number of hyperplanes that do not separate nor cut through either of the groups, and simplified using the assumptions that $\theta_1 = \frac{A_1}{2}$ and $A_1 = A_2$. 

Note that if we did not have the condition $k_{1,\theta_2} < 9(j+k_{1,\theta_1})$ in the sum (\ref{RF3 bound multinomial}) (that is, if we summed over all terms), the quantity would sum to 1 (this can easily be seen by the Multinomial Theorem). Finally, this means (\ref{RF3 bound multinomial}) is equivalent to (\ref{RF3 bound multinomial complement}), thereby completing the proof.
\end{proof}

\subsubsection{Proof of Corollary \ref{Corollary 1}}

\begin{proof}
We can  bound (\ref{RF3 bound multinomial complement}) from below by  bounding the excluded terms in the sum (i.e., those that satisfy $k_{1,\theta_2} \geq 9(j+k_{1,\theta_1})$) from above. One approach to this would be to count the number of terms satisfying $k_{1,\theta_2} \geq 9(j+k_{1,\theta_1})$ and bound them by their maximum.  Using basic combinatorics (see the appendix, Section \ref{app:comb}), 
that the number of terms satisfying $k_{1,\theta_2} \geq 9(j+k_{1,\theta_1})$ is given by
\begin{align} \label{multinomial count}
W_1 = \frac{1}{12} \left(\left\lfloor \frac{m}{10} \right\rfloor + 1\right) \left(\left\lfloor \frac{m}{10} \right\rfloor + 2\right) \left(150 \left\lfloor \frac{m}{10} \right\rfloor^2 - 10(4m + 1)\left\lfloor \frac{m}{10} \right\rfloor +3(m^2 + 3m + 2)\right) \sim m^4.
\end{align}
Then, the quantity (\ref{RF3 bound multinomial complement}) can be bounded below by
\begin{align} 
&1 - W_1 \max\left( \binom{m}{j,k_{1,\theta_1},k_{1,\theta_2},k_2,k} \left(\frac{A_{12}}{\pi}\right)^j \left(\frac{A_1}{2\pi} \right)^{k_{1,\theta_1}+k_{1,\theta_2}}   \left(\frac{A_1}{\pi} \right)^{k_2} \left(\frac{\pi-2A_1-A_{12}}{\pi}\right)^k \right) \notag\\
&= 1 - W_1 \max\left( \binom{m}{j,k_{1,\theta_1},k_{1,\theta_2},k_2,k} \left(\frac{1}{2} \right)^{k_{1,\theta_1}+k_{1,\theta_2}} \left(\frac{A_{12}}{\pi}\right)^j \left(\frac{A_1}{\pi} \right)^{k_{1,\theta_1}+k_{1,\theta_2}+k_2} \left(\frac{\pi-2A_1-A_{12}}{\pi}\right)^k \right), \label{RF3 bound multinomial estimate}
\end{align}
where the maximum is taken over all $j,k_{1,\theta_1},k_{1,\theta_2}, k_2, k = 0,\dots,m$ such that $k_{1,\theta_2} \geq 9(j+k_{1,\theta_1})$.
Ignoring the constraint  $k_{1,\theta_2} \geq 9(j+k_{1,\theta_1})$, we can upper bound the multinomial coefficient.

That is, assuming $\frac{m}{5} \in \mathbb{Z}$ for simplicity and applying Stirling's approximation for the factorial $n! \sim \sqrt{2\pi n} (\frac{n}{e})^n$, we get 
\begin{align}
\binom{m}{j,k_{1,\theta_1},k_{1,\theta_2},k_2,k}  &\leq \frac{m!}{[(\frac{m}{5})!]^5} \notag\\
&\sim \frac{\sqrt{2\pi m} (\frac{m}{e})^m}{[\sqrt{2\pi \frac{m}{5}} (\frac{m}{5e})^{m/5}]^5} \notag\\
&= \frac{5^{m + 5/2}}{(2\pi m)^2}. \label{multinomial bound}
\end{align}
Since we are assuming $A_{12}$ is larger than $A_1$ and $\pi-2A_1-A_{12}$, the strategy is to take $j$ to be as large as possible while satisfying $k_{1,\theta_2} \geq 9j$ and $j + k_{1,\theta_2} = m$. Since $k_{1,\theta_2} \geq 9j$, we have $j + 9j \leq m$ which implies $j \leq \frac{m}{10}$. So, we take $j= \frac{m}{10}$, $k_{1,\theta_2} = \frac{9m}{10}$, and $k_{1,\theta_1} = k_2 = k = 0$.  
Then
\begin{align}
 \left(\frac{1}{2} \right)^{k_{1,\theta_1}+k_{1,\theta_2}} \left(\frac{A_{12}}{\pi}\right)^j \left(\frac{A_1}{\pi} \right)^{k_{1,\theta_1}+k_{1,\theta_2}+k_2} \left(\frac{\pi-2A_1-A_{12}}{\pi}\right)^k &\leq \left(\frac{1}{2} \right)^{9m/10} \left(\frac{A_{12}}{\pi} \right)^{m/10} \left(\frac{A_1}{\pi} \right)^{9m/10} \notag\\
 &= \left(\frac{1}{2^9} \frac{A_{12}}{\pi} \left(\frac{A_1}{\pi}\right)^9 \right)^{m/10}.\label{rest bound}
\end{align}
Combining (\ref{RF3 bound multinomial estimate}) with the bounds given in (\ref{multinomial bound}) and (\ref{rest bound}), we have
\begin{align}
&\geq 1 - W_1  \frac{5^{m + 5/2}}{(2\pi m)^2}  \left(\frac{1}{2^9} \frac{A_{12}}{\pi} \left(\frac{A_1}{\pi}\right)^9 \right)^{m/10} \notag\\
&\sim 1- m^4  \frac{5^{m + 5/2}}{(2\pi m)^2}  \left(\frac{1}{2^9} \frac{A_{12}}{\pi} \left(\frac{A_1}{\pi}\right)^9 \right)^{m/10} \notag\\
&= 1- m^2  \frac{5^{5/2}}{(2\pi)^2}  \left(5^{10} \frac{1}{2^9} \frac{A_{12}}{\pi} \left(\frac{A_1}{\pi}\right)^9 \right)^{m/10}.
\end{align}
For the above to tend to 1 as $m\rightarrow \infty$, we need $ \frac{5^{10}}{2^9}  \frac{A_{12}}{\pi} \left(\frac{A_1}{\pi}\right)^9 < 1$. This is equivalent to $A_{12} \left(\frac{A_1}{2}\right)^9 < \frac{\pi^{10}}{5^{10}}$, which implies $A_{12} \theta_1^9 < \left(\frac{\pi}{5}\right)^{10} = \frac{\pi}{5}\left(\frac{\pi}{5}\right)^{9}$. Note that if $\theta_1 = \frac{\pi}{5}$, then $A_1 = A_2 = 2\theta_1 = \frac{2\pi}{5}$. Then $A_{12}$ could be at most $\frac{\pi}{5}$. But, this can't be because we have assumed $A_{12} \geq A_1$. Thus, we must have $\theta_1 < \frac{\pi}{5}$. In fact, $\theta_1 = \frac{\pi}{6}$ is the largest possible, in which case $A_{12} = A_1 = A_2 = \frac{\pi}{3}$. If $ \theta_1 = \frac{\pi}{6}$, then $A_{12} \theta_1^9 < \frac{\pi}{5}\left(\frac{\pi}{5}\right)^{9}$ becomes $A_{12} < \frac{\pi}{5} \left(\frac{6}{5} \right)^9 \approx 3.24$. Therefore, since we are already assuming $A_{12} + 2A_1 \leq \pi$, this is essentially no further restriction on $A_{12}$, and the same would be true for all $\theta_1 \leq \frac{\pi}{6}$. This completes the proof.
\end{proof}

\subsubsection{Proof of Corollary \ref{Corollary 2}}

\begin{proof}
Consider (\ref{RF3 bound multinomial complement})  and set $j' = j + k_{1,\theta_1}$ and $r = k_2 + k$. Then we view (\ref{RF3 bound multinomial complement}) as a probability equivalent to 
\begin{align}\label{equivalent probability}
1 - \underset{j'+k_{1,\theta_2}+r = m, \,\, k_{1,\theta_2} \geq 9j'}{\sum_{j'=0}^{2m} \sum_{k_{1,\theta_2}=0}^m \sum_{r=0}^{2m}} \binom{m}{k_{1,\theta_2},j',r} \left(\frac{A_{12}+\frac{A_1}{2}}{\pi} \right)^{j'} \left(\frac{A_1}{2\pi} \right)^{k_{1,\theta_2}} \left(\frac{\pi-A_1-A_{12}}{\pi} \right)^{r}.
\end{align}
Note that multinomial coefficients are maximized when the parameters all attain the same value. Thus, the multinomial term above is maximized when $  k_{1,\theta_2}$, $j'$ and $r$ are all as close to one another as possible. Thus, given the additional constraint that $k_{1,\theta_2} \geq 9j'$, 
the multinomial term is maximized when $k_{1,\theta_2}=\frac{9m}{19}$, $j' = \frac{m}{19}$, and $r = \frac{9m}{19}$ (possibly with ceilings/floors as necessary if $m$ is not a multiple of 19), (see the appendix, Section \ref{app:facts}, for a quick explanation),
which means
\begin{align}
\binom{m}{k_{1,\theta_2},j',r} &\leq \frac{m!}{(\frac{9m}{19})!(\frac{m}{19})!(\frac{9m}{19})!} \label{facts} \\
&\sim \frac{\sqrt{2\pi m}(\frac{m}{e})^m}{2\pi \frac{9m}{19} (\frac{9m}{19e})^{18m/19} \sqrt{2\pi \frac{m}{19}}(\frac{m}{19e})^{m/19} }\label{Stirling step} \\
&= \frac{19\sqrt{19}}{18\pi m} \left( (\frac{19}{9})^{18/19 19^{1/19}} \right)^m \notag \\
&\approx \frac{19\sqrt{19}}{18\pi m} 2.37^m, \label{trinomial bound}
\end{align}
where (\ref{Stirling step}) follows from Stirling's approximation for the factorial (and we use the notation $\sim$ to denote asymptotic equivalence, i.e. that two quantities have a ratio that tends to 1 as the parameter size grows). 

Now assume $A_{12} + \frac{3}{4}A_1 \leq \frac{\pi}{2}$, which implies $\pi-A_1-A_{12} \geq A_{12} + \frac{A_1}{2}$. Note also that $\pi-A_1 - A_{12} \geq A_1$ since it is assumed that $\pi-2A_1 - A_{12}\geq 0$. Therefore, we can lower bound (\ref{equivalent probability}) by
\begin{align}\label{corollary 2 bound}
1 - W_2\frac{19\sqrt{19}}{18\pi m} 2.37^m \left(\frac{\pi-A_1-A_{12}}{\pi} \right)^m,
\end{align}
where $W_2$ is the number of terms in the summation in (\ref{equivalent probability}), and is given by
\begin{align}
W_2= \frac{1}{6} \left(\left\lfloor \frac{m}{10} \right\rfloor + 1\right) \left(100 \left\lfloor \frac{m}{10} \right\rfloor^2 + (5 - 30m)\left\lfloor \frac{m}{10} \right\rfloor +3(m^2 + 3m + 2)\right) \sim m^3.
\end{align}
Thus, (\ref{corollary 2 bound}) goes to 1 as $m\rightarrow \infty$ when $2.37\left( \frac{\pi-A_1-A_{12}}{\pi}\right) <1 $, which holds if $A_1+A_{12} > 0.58\pi$.
\end{proof}

\section{Discussion and Conclusion}\label{sec::conclude}
In this work, we have presented a supervised classification algorithm that operates on binary, or one-bit, data. Along with encouraging numerical experiments, we have also included a theoretical analysis for a simple case. We believe our framework and analysis approach is relevant to analyzing similar, layered-type algorithms. 
Future directions of this work include the use of dithers for more complicated data geometries, as well as a generalized theory for high dimensional data belonging to many classes and utilizing multiple layers within the algorithm.

\appendix

\section{Elementary Computations}

\subsection{Derivation of \eqref{helloerf} and \eqref{helloerf2}}\label{app:erf}
The expected values above are related to the moment generating function of squares of uniform random variables. Let $Y= U^2$ where $U\sim U(a,b)$. Then the pdf of Y is given by
\begin{align}
f_Y(y) = \begin{cases} 
      \frac{1}{\sqrt{y}(b-a)} & a^2 \leq y \leq b^2 \\
      0 & \mbox{otherwise}.
   \end{cases}
\end{align}
Then,
\begin{align}
\mathbb{E}\left(e^{\frac{\theta}{c}U^2}\right) &= \mathbb{E}\left(e^{\frac{\theta}{c}Y}\right) \\
&= \int_{-\infty}^{\infty} e^{\frac{\theta}{c}y} f_Y(y) \,\, dy \\
&= \frac{1}{b-a} \int_{a^2}^{b^2} \frac{1}{\sqrt{y}}e^{\frac{\theta}{c}y} \,\, dy \\
&= \frac{2}{b-a} \int_a^b e^{\frac{\theta}{c}x^2} \,\, dx \\
&= \frac{\sqrt{\pi}(\mbox{erfi}(b\sqrt{\frac{\theta}{c}})-\mbox{erfi}(a\sqrt{\frac{\theta}{c}}))}{\sqrt{\frac{\theta}{c}}(b-a)}, \label{erfi mgf}
\end{align}
where we have used the substitution $x=\sqrt{y}$. 
Similarly,
\begin{align}
\mathbb{E}\left(e^{-\frac{\theta}{c}U^2}\right) &= \frac{\sqrt{\pi}(\mbox{erf}(b\sqrt{\frac{\theta}{c}})-\mbox{erf}(a\sqrt{\frac{\theta}{c}}))}{\sqrt{\frac{\theta}{c}}(b-a)}, \label{erf mgf}
\end{align}
where $\mbox{erf}(x) = \frac{2}{\sqrt{\pi}} \int_0^x e^{-t^2} \,\, dt$ is the error function and $\mbox{erfi}(x) = -i\mbox{erf}(ix) = \frac{2}{\sqrt{\pi}}\int_0^x e^{t^2} \,\, dt$ is the imaginary error function. Then  \eqref{helloerf} and \eqref{helloerf2} hold by observing that $\mbox{erf}(0) = \mbox{erfi}(0) = 0$.

\subsection{Derivation of \eqref{multinomial count}}\label{app:comb}

Suppose we have $M$ objects that must be divided into 5 boxes (for us, the boxes are the 5 different types of hyperplanes). Let $n_i$ denote the number of objects put into box $i$. Recall that in general, $M$ objects can be divided into $k$ boxes $\binom{M+k-1}{k-1}$ ways.

How many arrangements satisfy $n_1 \geq 9(n_2 + n_3)$? To simplify, let $n$ denote the total number of objects in boxes 2 and 3 (that is, $n = n_2 + n_3$). Then, we want to know how many arrangements satisfy $n_1 \geq 9n$? 

If $n=0$, then $n_1 \geq 9n$ is satisfied no matter how many objects are in box 1. So, this reduces to the number of ways to arrange $M$ objects into 3 boxes, which is given by $\binom{M+2}{2}$.

Suppose $n=1$. For $n_1 \geq 9n$ to be true, we must at least reserve 9 objects in box 1. Then $M-10$ objects remain to be placed in 3 boxes, which can be done in $\binom{(M-10)+2}{2}$ ways. But, there are 2 ways for $n=1$, either $n_2=1$ or $n_3 = 1$, so we must multiply this by 2. Thus, $\binom{(M-10)+2}{2} \times 2$ arrangements satisfy  $n_1 \geq 9n$.

Continuing in this way, in general for a given $n$, there are $\binom{M-10n+2}{2} \times (n+1)$ arrangements that satisfy $n_1 \geq 9n$. There are $n+1$ ways to arrange the objects in boxes 2 and 3, and $\binom{M-10n+2}{2}$ ways to arrange the remaining objects after $9n$ have been reserved in box 1. 

Therefore, the total number of arrangements that satisfy $n_1 \geq 9n$ is given by
\begin{align}
\sum_{n=0}^{\lfloor \frac{M}{10} \rfloor} \binom{M-10n+2}{2} \times (n+1).
\end{align}
To see the upper limit of the sum above, note that we must have $M-10n+2 \geq 2$, which means $n \leq \frac{M}{10}$. Since $n$ must be an integer, we take $n \leq \lfloor \frac{M}{10} \rfloor$. After some heavy algebra (i.e. using software!), one can express this sum as:

\begin{align}
W = \frac{1}{12} \left(\left\lfloor \frac{M}{10} \right\rfloor + 1\right) \left(\left\lfloor \frac{M}{10} \right\rfloor + 2\right) \left(150 \left\lfloor \frac{M}{10} \right\rfloor^2 - 10(4M + 1)\left\lfloor \frac{M}{10} \right\rfloor +3(M^2 + 3M + 2)\right) \sim M^4.
\end{align}

\subsection{Derivation of \eqref{facts}}\label{app:facts}
Suppose we want to maximize (over the choices of $a,b,c$) a trinomial $\frac{m!}{a!b!c!}$ subject to $a+b+c=m$ and $a>9b$. Since $m$ is fixed, this is equivalent to choosing $a,b,c$ so as to minimize $a!b!c!$ subject to these constraints. First, fix $c$ and consider optimizing $a$ and $b$ subject to $a+b = m-c =: k$ and $a>9b$ in order to minimize $a!b!$. For convenience, suppose $k$ is a multiple of $10$. We claim the optimal choice is to set $a=9b$ (i.e. $a = \frac{9}{10}k$ and $b=\frac{1}{10}k$). Write $a = 9b + x$ where $x$ must be some non-negative integer in order to satisfy the constraint. We then wish to compare $(9b)!b!$ to $(9b+x)!(b-x)!$, since the sum of $a$ and $b$ must be fixed. One readily observes that:
$$
(9b+x)!(b-x)! = \frac{(9b+x)(9b+x-1)\cdots(9b+1)}{b(b-1)\cdots(b-x+1)}\cdot (9b)!b! \geq \frac{9b\cdot 9b\cdots 9b}{b\cdot b\cdots b}\cdot (9b)!b! = 9^x\cdot (9b)!b!.
$$
Thus, we only increase the product $a!b!$ when $a>9b$, so the optimal choice is when $a=9b$. This holds for any choice of $c$.  A similar argument shows that optimizing $b$ and $c$ subject to $9b + b + c = m$ to minimize $(9b)!b!c!$ results in the choice that $c=9b$.  Therefore, one desires that $a=c=9b$ and $a+b+c=m$, which means $a=c=\frac{9}{19}m$ and $b=\frac{1}{19}m$.

\bibliographystyle{plain}
\bibliography{bib}

\begin{thebibliography}{10}

\bibitem{achlioptas2003database}
Dimitris Achlioptas.
\newblock Database-friendly random projections: Johnson-lindenstrauss with
  binary coins.
\newblock {\em Journal of computer and System Sciences}, 66(4):671--687, 2003.

\bibitem{ailon2006approximate}
Nir Ailon and Bernard Chazelle.
\newblock Approximate nearest neighbors and the fast johnson-lindenstrauss
  transform.
\newblock In {\em Proceedings of the thirty-eighth annual ACM symposium on
  Theory of computing}, pages 557--563. ACM, 2006.

\bibitem{andrew2000introduction}
Alex~M Andrew.
\newblock An introduction to support vector machines and other kernel-based
  learning methods by nello christianini and john shawe-taylor, cambridge
  university press, cambridge, 2000, xiii+ 189 pp., isbn 0-521-78019-5
  (hbk,{\pounds} 27.50)., 2000.

\bibitem{aziz1996overview}
Pervez~M Aziz, Henrik~V Sorensen, and J~Vn~der Spiegel.
\newblock An overview of sigma-delta converters.
\newblock {\em IEEE signal processing magazine}, 13(1):61--84, 1996.

\bibitem{baraniuk2006johnson}
Richard Baraniuk, Mark Davenport, Ronald DeVore, and Michael Wakin.
\newblock The johnson-lindenstrauss lemma meets compressed sensing.
\newblock {\em preprint}, 100(1):0, 2006.

\bibitem{BoufoB_1Bit}
P.~Boufounos and R.~Baraniuk.
\newblock 1-bit compressive sensing.
\newblock In {\em Proc. IEEE Conf. Inform. Science and Systems (CISS)},
  Princeton, NJ, March 2008.

\bibitem{CHH07b}
Deng Cai, Xiaofei He, and Jiawei Han.
\newblock Spectral regression for efficient regularized subspace learning.
\newblock In {\em Proc. Int. Conf. Computer Vision (ICCV'07)}, 2007.

\bibitem{CHHZ06}
Deng Cai, Xiaofei He, Jiawei Han, and Hong-Jiang Zhang.
\newblock Orthogonal laplacianfaces for face recognition.
\newblock {\em IEEE Transactions on Image Processing}, 15(11):3608--3614, 2006.

\bibitem{CHHH07}
Deng Cai, Xiaofei He, Yuxiao Hu, Jiawei Han, and Thomas Huang.
\newblock Learning a spatially smooth subspace for face recognition.
\newblock In {\em Proc. IEEE Conf. Computer Vision and Pattern Recognition
  Machine Learning (CVPR'07)}, 2007.

\bibitem{CandeRT_Robust}
E.~Cand\`{e}s, J.~Romberg, and T.~Tao.
\newblock Robust uncertainty principles: {E}xact signal reconstruction from
  highly incomplete frequency information.
\newblock {\em IEEE Trans. Inform. Theory}, 52(2):489--509, 2006.

\bibitem{CandeRT_Stable}
E.~Cand\`{e}s, J.~Romberg, and T.~Tao.
\newblock Stable signal recovery from incomplete and inaccurate measurements.
\newblock {\em Comm. Pure Appl. Math.}, 59(8):1207--1223, 2006.

\bibitem{candy1962oversampling}
James~C Candy and Gabor~C Temes.
\newblock {\em Oversampling delta-sigma data converters: theory, design, and
  simulation}.
\newblock University of Texas Press, 1962.

\bibitem{choromanska2016binary}
Anna Choromanska, Krzysztof Choromanski, Mariusz Bojarski, Tony Jebara, Sanjiv
  Kumar, and Yann LeCun.
\newblock Binary embeddings with structured hashed projections.
\newblock In {\em Proceedings of The 33rd International Conference on Machine
  Learning}, pages 344--353, 2016.

\bibitem{CristS_Introduction}
N.~Christianini and J.~Shawe-Taylor.
\newblock {\em An Introduction to Support Vector Machines and Other
  Kernel-Based Learning Methods}.
\newblock Cambridge University Press, Cambridge, England, 2000.

\bibitem{dasgupta2003elementary}
Sanjoy Dasgupta and Anupam Gupta.
\newblock An elementary proof of a theorem of johnson and lindenstrauss.
\newblock {\em Random Structures \& Algorithms}, 22(1):60--65, 2003.

\bibitem{dirksen2016fast}
Sjoerd Dirksen and Alexander Stollenwerk.
\newblock Fast binary embeddings with gaussian circulant matrices: improved
  bounds.
\newblock {\em arXiv preprint arXiv:1608.06498}, 2016.

\bibitem{Donoh_Compressed}
D.~Donoho.
\newblock Compressed sensing.
\newblock {\em IEEE Trans. Inform. Theory}, 52(4):1289--1306, 2006.

\bibitem{fang2014sparse}
Jun Fang, Yanning Shen, Hongbin Li, and Zhi Ren.
\newblock Sparse signal recovery from one-bit quantized data: An iterative
  reweighted algorithm.
\newblock {\em Signal Processing}, 102:201--206, 2014.

\bibitem{gong2013iterative}
Yunchao Gong, Svetlana Lazebnik, Albert Gordo, and Florent Perronnin.
\newblock Iterative quantization: A procrustean approach to learning binary
  codes for large-scale image retrieval.
\newblock {\em IEEE Transactions on Pattern Analysis and Machine Intelligence},
  35(12):2916--2929, 2013.

\bibitem{gopi2013one}
Sivakant Gopi, Praneeth Netrapalli, Prateek Jain, and Aditya~V Nori.
\newblock One-bit compressed sensing: {P}rovable support and vector recovery.
\newblock In {\em ICML (3)}, pages 154--162, 2013.

\bibitem{gupta2010sample}
Ankit Gupta, Robert Nowak, and Benjamin Recht.
\newblock Sample complexity for 1-bit compressed sensing and sparse
  classification.
\newblock In {\em Information Theory Proceedings (ISIT), 2010 IEEE
  International Symposium on}, pages 1553--1557. IEEE, 2010.

\bibitem{hahn2014adaptive}
Jurgen Hahn, Simon Rosenkranz, and Abdelhak~M Zoubir.
\newblock Adaptive compressed classification for hyperspectral imagery.
\newblock In {\em Acoustics, Speech and Signal Processing (ICASSP), 2014 IEEE
  International Conference on}, pages 1020--1024. IEEE, 2014.

\bibitem{HYHNZ05}
Xiaofei He, Shuicheng Yan, Yuxiao Hu, Partha Niyogi, and Hong-Jiang Zhang.
\newblock Face recognition using laplacianfaces.
\newblock {\em IEEE Trans. Pattern Anal. Mach. Intelligence}, 27(3):328--340,
  2005.

\bibitem{hearst1998support}
Marti~A. Hearst, Susan~T Dumais, Edgar Osuna, John Platt, and Bernhard
  Scholkopf.
\newblock Support vector machines.
\newblock {\em IEEE Intelligent Systems and their Applications}, 13(4):18--28,
  1998.

\bibitem{hunter2010compressive}
Blake Hunter, Thomas Strohmer, Theodore~E Simos, George Psihoyios, and
  Ch~Tsitouras.
\newblock Compressive spectral clustering.
\newblock In {\em AIP Conference Proceedings}, volume 1281, pages 1720--1722.
  AIP, 2010.

\bibitem{jacques2013quantized}
Laurent Jacques, K{\'e}vin Degraux, and Christophe De~Vleeschouwer.
\newblock Quantized iterative hard thresholding: Bridging 1-bit and
  high-resolution quantized compressed sensing.
\newblock {\em arXiv preprint arXiv:1305.1786}, 2013.

\bibitem{JacquLBB_Robust}
Laurent Jacques, Jason Laska, Petros Boufounos, and Richard Baraniuk.
\newblock {Robust 1-bit compressive sensing via binary stable embeddings of
  sparse vectors}.
\newblock {\em IEEE Trans. Inform. Theory}, 59(4):2082--2102, 2013.

\bibitem{joachims1998text}
Thorsten Joachims.
\newblock Text categorization with support vector machines: Learning with many
  relevant features.
\newblock {\em Machine learning: ECML-98}, pages 137--142, 1998.

\bibitem{JohnsL_Extensions}
W.~Johnson and J.~Lindenstrauss.
\newblock Extensions of {L}ipschitz mappings into a {H}ilbert space.
\newblock In {\em Proc. Conf. Modern Anal. and Prob.}, New Haven, CT, June
  1982.

\bibitem{krahmer2011new}
Felix Krahmer and Rachel Ward.
\newblock New and improved johnson--lindenstrauss embeddings via the restricted
  isometry property.
\newblock {\em SIAM Journal on Mathematical Analysis}, 43(3):1269--1281, 2011.

\bibitem{krizhevsky2012imagenet}
Alex Krizhevsky, Ilya Sutskever, and Geoffrey~E Hinton.
\newblock Imagenet classification with deep convolutional neural networks.
\newblock In {\em Advances in neural information processing systems}, pages
  1097--1105, 2012.

\bibitem{LaskaWYB_Trust}
Jason~N Laska, Zaiwen Wen, Wotao Yin, and Richard~G Baraniuk.
\newblock {Trust, but verify: Fast and accurate signal recovery from 1-bit
  compressive measurements}.
\newblock {\em IEEE Trans. Signal Processing}, 59(11):5289--5301, 2011.

\bibitem{MNIST}
Y.~LeCun.
\newblock The mnist database of handwritten digits.
\newblock \url{http://yann.lecun.com/exdb/mnist/}.

\bibitem{PlanV_One}
Y.~Plan and R.~Vershynin.
\newblock One-bit compressed sensing by linear programming.
\newblock {\em Communications on Pure and Applied Mathematics},
  66(8):1275--1297, 2013.

\bibitem{PlanV_Robust}
Y.~Plan and R.~Vershynin.
\newblock Robust 1-bit compressed sensing and sparse logistic regression: {A}
  convex programming approach.
\newblock {\em IEEE Trans. Inform. Theory}, 59(1):482--494, 2013.

\bibitem{PlanV_Dimension}
Y.~Plan and R.~Vershynin.
\newblock Dimension reduction by random hyperplane tessellations.
\newblock {\em Discrete \& Computational Geometry}, 51(2):438--461, 2014.

\bibitem{russakovsky2015imagenet}
Olga Russakovsky, Jia Deng, Hao Su, Jonathan Krause, Sanjeev Satheesh, Sean Ma,
  Zhiheng Huang, Andrej Karpathy, Aditya Khosla, Michael Bernstein, et~al.
\newblock Imagenet large scale visual recognition challenge.
\newblock {\em International Journal of Computer Vision}, 115(3):211--252,
  2015.

\bibitem{simonyan2014very}
Karen Simonyan and Andrew Zisserman.
\newblock Very deep convolutional networks for large-scale image recognition.
\newblock {\em arXiv preprint arXiv:1409.1556}, 2014.

\bibitem{steinwart2008support}
Ingo Steinwart and Andreas Christmann.
\newblock {\em Support vector machines}.
\newblock Springer Science \& Business Media, 2008.

\bibitem{szegedy2015going}
Christian Szegedy, Wei Liu, Yangqing Jia, Pierre Sermanet, Scott Reed, Dragomir
  Anguelov, Dumitru Erhan, Vincent Vanhoucke, and Andrew Rabinovich.
\newblock Going deeper with convolutions.
\newblock In {\em Proceedings of the IEEE Conference on Computer Vision and
  Pattern Recognition}, pages 1--9, 2015.

\bibitem{yan2012robust}
Ming Yan, Yi~Yang, and Stanley Osher.
\newblock Robust 1-bit compressive sensing using adaptive outlier pursuit.
\newblock {\em IEEE Trans. Signal Processing}, 60(7):3868--3875, 2012.

\bibitem{price2015binary}
Xinyang Yi, Constantine Caravans, and Eric Price.
\newblock Binary embedding: Fundamental limits and fast algorithm.
\newblock 2015.

\bibitem{yu2014circulant}
Felix~X Yu, Sanjiv Kumar, Yunchao Gong, and Shih-Fu Chang.
\newblock Circulant binary embedding.
\newblock In {\em International conference on machine learning}, volume~6,
  page~7, 2014.

\end{thebibliography}

\end{document}